\documentclass[acmtog]{acmart}

\AtBeginDocument{%
  \providecommand\BibTeX{{%
    \normalfont B\kern-0.5em{\scshape i\kern-0.25em b}\kern-0.8em\TeX}}}

\usepackage{colortbl}
\usepackage{siunitx}
\usepackage{xfrac}
\usepackage{accents}
\usepackage{paralist}

\usepackage{booktabs} %
\usepackage{xspace}
\usepackage{array}
\usepackage{siunitx}
\usepackage{hyperref}
\usepackage{dblfloatfix} %
\newcolumntype{L}[1]{>{\raggedright\let\newline\\\arraybackslash\hspace{0pt}}m{#1}}
\newcolumntype{C}[1]{>{\centering\let\newline\\\arraybackslash\hspace{0pt}}m{#1}}
\newcolumntype{R}[1]{>{\raggedleft\let\newline\\\arraybackslash\hspace{0pt}}m{#1}}

\newcommand{\sect}[1]{Section~\ref{#1}}

\newcommand{\fig}[1]{Figure~\ref{#1}}
\newcommand{\tbl}[1]{Table~\ref{#1}}

\newcommand{\degree}{\ensuremath{^\circ}\xspace}
\newcommand{\ignore}[1]{}

\def\around{{\raise.17ex\hbox{$\scriptstyle\mathtt{\sim}$}}}

\makeatletter
\DeclareRobustCommand\onedot{\futurelet\@let@token\@onedot}
\def\@onedot{\ifx\@let@token.\else.\null\fi\xspace}

\def\eg{e.g\onedot} 
\def\ie{i.e\onedot} 
 
\def\etc{etc\onedot} \def\vs{vs\onedot}
\def\wrt{w.r.t\onedot} 

\def\aka{a.k.a\onedot}
\makeatother

\definecolor{MyLightGreen}{rgb}{0.92,1,0.95}
\definecolor{MyLightBlue}{rgb}{0.92,0.95,1}
\definecolor{MyDarkBlue}{rgb}{0,0.08,1}
\definecolor{MyDarkGreen}{rgb}{0.02,0.6,0.02}
\definecolor{MyDarkCyan}{rgb}{0,0.7,0.7}
\definecolor{MyDarkOrange}{rgb}{0.40,0.2,0.02}
\definecolor{MyPurple}{RGB}{111,0,255}
\definecolor{MyRed}{rgb}{1.0,0.0,0.0}
\definecolor{MyDarkRed}{rgb}{0.8,0,0}
\definecolor{MyGold}{rgb}{0.75,0.6,0.12}
\definecolor{MyDarkgray}{rgb}{0.66, 0.66, 0.66}

\def\to{$\,\to\,$}

\newcommand{\model}{NLT\xspace}
\newcommand{\Model}{NLT\xspace}

\newcommand{\nltfull}{Neural Light Transport (NLT)\xspace}
\newcommand{\nlt}{NLT\xspace}

\newcommand{\obspath}{observation paths\xspace}
\newcommand{\querypath}{query path\xspace}

\newcommand{\myparagraph}[1]{\vspace{-4pt}\paragraph{#1}}

\newcommand{\conv}{\ensuremath{\operatorname{conv}}}
\newcommand{\upconv}{\ensuremath{\operatorname{conv}^{\mathrm{T}}}}
\newcommand{\append}{\ensuremath{\operatorname{append}}\,}
\newcommand{\mean}{\ensuremath{\operatorname{mean}}}

\newcommand{\light}{\boldsymbol{\omega}}
\newcommand{\viewdir}{\light_o}
\newcommand{\outdir}{\light_i}

\newcommand{\losfun}{\mathcal{L_I}}
\newcommand{\losfunp}{\mathcal{L_P}}

\newcommand{\materialediting}[1]{}

\newcommand{\rev}[1]{#1} 

\setcopyright{acmlicensed}
\acmJournal{TOG}
\acmYear{2020}
\acmVolume{40}
\acmNumber{1}
\acmMonth{12}
\acmPrice{15.00}
\acmDOI{10.1145/3446328}
\setcitestyle{square,nosort}

\begin{document}

\title{Neural Light Transport for Relighting and View Synthesis}

\author{Xiuming Zhang}
\email{xiuming@csail.mit.edu}
\orcid{0000-0002-4326-727X}
\affiliation{%
  \institution{Massachusetts Institute of Technology}
}

\author{Sean Fanello}
\author{Yun-Ta Tsai}
\affiliation{%
  \institution{Google}
}

\author{Tiancheng Sun}
\affiliation{%
  \institution{University of California, San Diego}
}

\author{Tianfan Xue}
\author{Rohit Pandey}
\author{Sergio Orts-Escolano}
\author{Philip Davidson}
\author{Christoph Rhemann}
\author{Paul Debevec}
\author{Jonathan T.\ Barron}
\affiliation{%
  \institution{Google}
}

\author{Ravi Ramamoorthi}
\affiliation{%
  \institution{University of California, San Diego}
}

\author{William T.\ Freeman}
\email{billf@mit.edu}
\affiliation{%
  \institution{Massachusetts Institute of Technology \& Google}
}

\renewcommand{\shortauthors}{Zhang, Fanello, Tsai, Sun, Xue, Pandey, Orts-Escolano, Davidson, Rhemann, Debevec, Barron, Ramamoorthi, and Freeman}

\begin{abstract}
The light transport (LT) of a scene describes how it appears under different lighting conditions from different viewing directions, and complete knowledge of a scene's LT enables the synthesis of novel views under arbitrary lighting.
In this paper, we focus on image-based LT acquisition, primarily for human bodies within a light stage setup.
We propose a semi-parametric approach for learning a neural representation of the LT that is embedded in a texture atlas of known but possibly rough geometry. We model all non-diffuse and global LT as residuals added to a physically-based diffuse base rendering.
In particular, we show how to fuse previously seen observations of illuminants and views to synthesize a new image of the same scene under a desired lighting condition from a chosen viewpoint.
This strategy allows the network to learn complex material effects (such as subsurface scattering) and global illumination (such as diffuse interreflection), while guaranteeing the physical correctness of the diffuse LT (such as hard shadows).
With this learned LT, one can relight the scene photorealistically with a directional light or an HDRI map, synthesize novel views with view-dependent effects, or do both simultaneously, all in a unified framework using a set of sparse observations. Qualitative and quantitative experiments demonstrate that our Neural Light Transport (\model) outperforms state-of-the-art solutions for relighting and view synthesis, without requiring separate treatments for both problems that prior work requires. The code and data are available at \href{http://nlt.csail.mit.edu}{\color{blue}{http://nlt.csail.mit.edu}}. %
\end{abstract} 
\begin{teaserfigure}
  \centering
   \includegraphics[width=\linewidth]{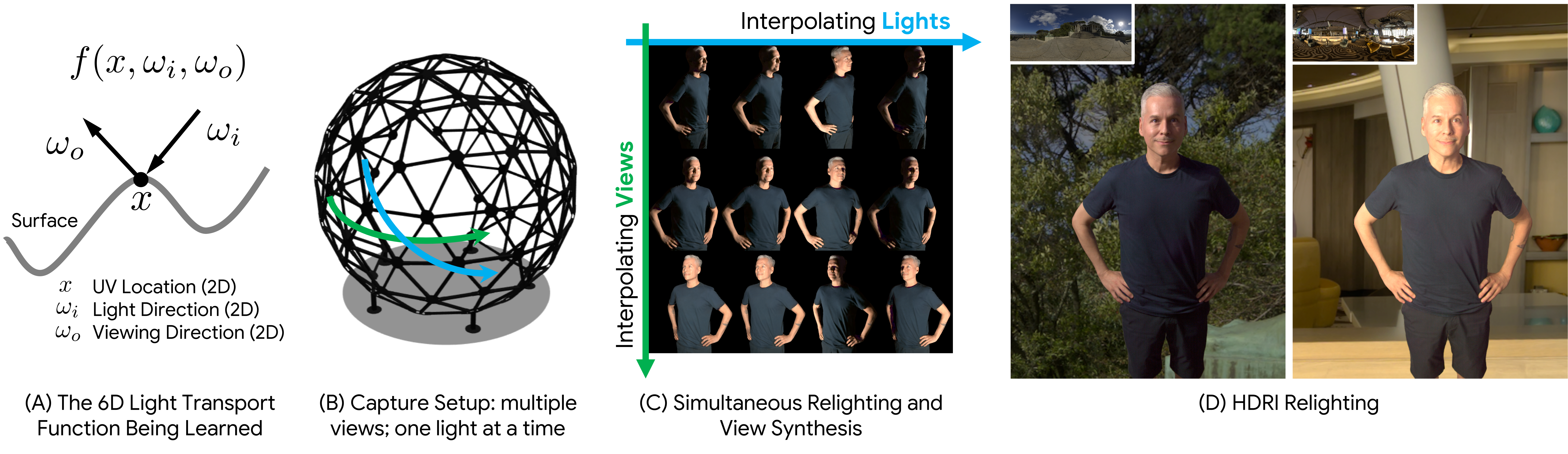}
\vspace{-.7cm}
  \caption{
  (A) Neural Light Transport (\model) learns to interpolate the 6D light transport function of a surface as a function of the UV coordinate (2 DOFs), incident light direction (2 DOFs), and viewing direction (2 DOFs).
  (B) The subject is imaged from multiple viewpoints when lit by different directional lights; a geometry proxy is also captured using active sensors.
  (C) Querying the learned function at different light and/or viewing directions enables simultaneous relighting and view synthesis of this subject.
  (D) The relit renderings that \model produces can be combined according to HDRI maps to perform image-based relighting.
    }
  \label{fig:teaser}
\end{teaserfigure} 

\maketitle

\section{Introduction}
The light transport (LT) of a scene models how light interacts with objects in the scene to produce an observed image. The process by which geometry and material properties of the scene interact with global illumination to result in an image is a complicated but well-understood consequence of physics~\cite{pharr}. Much progress in computer graphics has been through the development of more expressive and more efficient mappings from a scene model (geometry, materials, and lighting) to an image. In contrast, \emph{inverting} this process is ill-posed and therefore more difficult: acquiring the LT of a scene from images of that scene requires untangling the myriad interconnected effects of occlusion, shading, shadowing, interreflections, scattering, \etc.
Solving this task of inferring aspects of LT from images is an active research area, and even partial solutions have significant practical uses such as phototourism~\cite{snavely2006photo}, telepresence~\cite{holoportation}, storytelling \cite{aria}, and special effects~\cite{debevec2012light}. A less obvious, but equally important application of inferring LT from images consists of generating groundtruth data for machine learning tasks: many works rely on high-quality renderings of relit subjects under arbitrary lighting conditions and from multiple viewpoints to perform relighting \cite{meka_deep_2019,sun_single_2019}, view synthesis \cite{simply_lookingood}, re-enacting \cite{kim2018deep}, and alpha matting \cite{BMSengupta20}.

Previous work has shown that it is possible to construct a light stage~\cite{debevec_acquiring_2000}, plenoptic camera~\cite{lightfield}, or gantry~\cite{MurrayColeman_Smith} that directly captures a subset of the LT function and thereby enables the image-based rendering thereof. These techniques are widely used in film productions and within the research community. However, these systems can only provide \textit{sparse} sampling of the LT limited to the number of LEDs ($\sim$\! 300 on a spherical dome) and the number of cameras ($\sim$\! 50-100 around the subject), resulting in the inability to produce photorealistic renderings outside the supported camera/light locations. Indeed, traditional image-based rendering approaches are usually designed for fixed viewpoints and are unable to synthesize unseen (novel) views under a desired illumination.

In this paper, we learn to interpolate the \textit{dense} LT function of a given scene from \textit{sparse} multi-view, One-Light-at-A-Time (OLAT) images acquired in a light stage~\cite{debevec_acquiring_2000}, through a semi-parametric technique that we dub \nltfull (\fig{fig:teaser}). Many prior works have addressed similar tasks (as will be discussed in \sect{sec:related}), with classic works tending to rely on physics to recover analytical and interpretable models, and recent works using neural networks to infer a more direct mapping from input images to an output image.

Traditional rendering methods often make simplifying assumptions when modeling geometry, BRDFs, or complex inter-object interactions in order to make the problem tractable. On the other hand, deep learning approaches can tolerate geometric and reflectance imperfections, but they often require many aspects of image formation (even those guaranteed by physics) be learned ``from scratch,'' which may necessitate a prohibitively large training set. \nlt is intended to straddle this divide: we construct a classical model of the subject being imaged (a mesh and a diffuse texture atlas per Lambertian reflectance), but then we embed a neural network within the parameterization provided by that classical model, construct the inputs and outputs of the model in ways that leverage domain knowledge of classical graphics techniques, and train that network to model all aspects of LT---including those not captured by a classical model. By leveraging a classical model this way, \nlt is able to learn an accurate model of the complicated LT function for a subject from a small training dataset of sparse observations.

A key novelty of \nlt is that our learned model is embedded within the texture atlas space of an existing geometric model of the subject, which provides a novel framework for \emph{simultaneous} relighting and view interpolation.
We express the 6D LT function (\fig{fig:teaser}) at each location on the surface of our geometric model as simply the output of a deep neural network, which works well (as neural networks are smooth and universal function approximators~\cite{hornik1991approximation}) and obviates the need for a complicated parameterization of spatially-varying reflectance.
We evaluate on joint relighting and view synthesis using sparse image observations of scanned human subjects within a light stage, and show state-of-the-art results as well as compelling practical applications.

In summary, our main contributions are:

$\bullet$ An end-to-end, semi-parametric method for learning to interpolate the 6D light transport function per-subject from real data using convolutional neural networks (\sect{sec:queryobs});

$\bullet$  A unified framework for simultaneous relighting and view synthesis by embedding networks into a parameterized texture atlas and leveraging as input a set of One-Light-at-A-Time (OLAT) images (\sect{sec:simul});

$\bullet$  A set of augmented texture-space inputs and a residual learning scheme on top of a physically accurate diffuse base, which together allow the network to easily learn non-diffuse, higher-order light transport effects including specular highlights, subsurface scattering, and global illumination (\sect{sec:uvbuffers} and \sect{sec:residual}).
\vspace{.1cm}

\noindent The proposed method allows for photorealistic free-viewpoint rendering under controllable lighting conditions, which not only is a key aspect in compelling user experiences in mixed reality and special effects, but can be applied to a variety of machine learning tasks that rely on photorealistic groundtruth data.

\section{Related Work}
\label{sec:related}

Our method addresses the problem of recovering a model of light transport from a sparse set of images of some subject, and then predicting novel images of that subject from unseen views and/or under unobserved illuminations. This is a broad problem statement that relates to and subsumes many tasks in graphics and vision.

\myparagraph{Single observation}
\rev{
The most sparse sampling is just a single image, from which one could attempt to infer a model (geometry, reflectance, and illumination) of the physical world that resulted in that image~\cite{Barrow1978}, usually via hand-crafted~\cite{barron_shape_2015,li2020multi} or learned priors~\cite{saxena2008make3d,eigen2014depth,sfsnetSengupta18,Zhengqin2,relightinghumans,kim2018deep,gardner_deep_nodate,legendre2019deeplight,alldieck2019tex2shape,wiles2020synsin,zhang2020portrait}. 
Though practical, the quality gap between what can be accomplished by single-image techniques and what has been demonstrated by multi-image techniques is significant. Indeed, none of these methods shows complex light transport effects such as specular highlights  or subsurface scattering~\cite{relightinghumans,kim2018deep}.
Moreover, these methods are usually limited to a single task, such as relighting \cite{relightinghumans,kim2018deep,sfsnetSengupta18} or viewpoint change \cite{alldieck2019tex2shape,wiles2020synsin,li2020multi}, and some support only a limited range of viewpoint change~\cite{kim2018deep,tewari2020stylerig}.
}

\myparagraph{Multiple views}
Multiview geometry techniques recover a textured 3D model that can be rendered using conventional graphics or photogrammetry techniques \cite{Hartley2003}, but have material and shading variation baked in, and do not enable relighting.
Image-based rendering techniques such as light fields~\cite{lightfield} or lumigraphs \cite{Gortler} can be used to directly sample and render the plenoptic function~\cite{adelson1991plenoptic}, but the accuracy of these techniques is limited by the density of sampled input images.
For unstructured inputs, reprojection-based methods~\cite{buehler2001unstructured,eisemann2008floating} assume the availability of a geometry proxy (so does our work), reproject nearby views to the query view, and perform image blending in that view. However, such methods rely heavily on the quality of the geometry proxy and cannot synthesize pixels that are not visible in the input views. 
A class-specific geometry prior (such as that of a human body~\cite{shysheya_textured_nodate}) can be used to increase the accuracy of a geometry proxy~\cite{carranza2003free}, but none of these methods enables relighting.

Recently, deep learning techniques have been used to synthesize new images from sparse sets of input images, usually by training neural networks to synthesize some intermediate geometric representation that is then projected into the desired image~\cite{zhou_stereo_2018,sitzmann_deepvoxels:_2019,sitzmann2019scene,SrinivasanCVPR2019,Flynn_2019_CVPR, mildenhall2019llff,mildenhall2020nerf,thies2020image}.
Some techniques even entirely replace the rendering process with a learned ``neural'' renderer \cite{thies_deferred_2019,lookinggood,simply_lookingood,neural_volumes,lombardi_deep_2018,tewari2020state}.
Despite effective, these methods generally do not attempt to explicitly model light transport and hence do not enable relighting---though they are often capable of preserving view-dependent effects for the fixed illumination condition under which the input images were acquired \cite{thies_deferred_2019,mildenhall2020nerf}.
Additionally, neural rendering often breaks ``backwards compatibility'' with existing graphics systems, while our approach infers images directly in texture space that can be re-sampled by conventional graphics software (\eg, Unity, Blender, \etc) to synthesize novel viewpoints. %
 Recently, \citet{chen2020neural} propose to learn relightable view synthesis from dense views (200 \vs 55 in this work) under image-based lighting; using spherical harmonics as the lighting representation, the work is unable to produce hard shadow caused by a directional light as in this work.

\myparagraph{Multiple illuminants}
\rev{
Similar to the multi-view task is the task of photometric stereo~\cite{photometric_stereo, basri2007photometric}  (as cameras function analogously to illuminants in some contexts~\cite{sen_dual_2005}):
repeatedly imaging a subject with a fixed camera but under different illuminations and then recovering the surface normals. However, most photometric stereo solutions assume Lambertian reflectance and do not support relighting with non-diffuse light transport.
More recently, \citet{ren_image_2015}, \citet{meka_deep_2019}, \citet{sun_single_2019}, and \citet{sun2020light} show that neural networks can be applied to relight a scene captured under multiple lighting conditions from a fixed viewpoint.
\citet{nestmeyerlearning} decompose an image into shaded albedo (so no cast shadow) and residuals, unlike this work that models cast shadow as part of a physically accurate diffuse base.
None of these works supports view synthesis.
\citet{xu_deep_2019} perform free-viewpoint relighting, but unlike our approach, they
require running the model of ~\citet{xu_deep_2018} as a second stage. %
}

\myparagraph{Multiple views and illuminants}
\rev{
\citet{garg2006symmetric} utilize the symmetry of illuminations and view directions to collect sparse samples of an 8D reflectance field, and reconstruct a complete field using a low-rank assumption.
Perhaps the most effective approach for addressing sparsity in light transport estimation is to circumvent this problem entirely, and densely sample whatever is needed to produce the desired renderings.
The landmark work of \citet{debevec_acquiring_2000} uses a light stage to acquire the full reflectance field of a subject by capturing a One-Light-at-A-Time (OLAT) scan of that subject, which can be used to relight the subject by linear combination according to some High-Dynamic-Range Imaging (HDRI) environment map.
Despite its excellent results, this approach lacks an explicit geometric model, so rendering is limited to a fixed set of viewpoints. Although this limitation has been partially addressed by the work of \citet{Ma:2007} that focuses on facial capture,
 a recent system of \citet{guo_relightables:_2019} builds a full volumetric relightable model using two spherical gradient illumination conditions~\cite{Fyffe:2009}. This system supports relighting and view synthesis, but assumes pre-defined BRDFs and therefore cannot synthesize more complex light transport effects present in real images.
 }
\citet{zickler06} also pose the problem of appearance synthesis as that of high-dimensional interpolation, but they use radial basis functions on smaller-scale data.

\vspace{0.5em}
Our work follows the convention of the nascent field of ``neural rendering'' \cite{thies_deferred_2019,neural_volumes,lombardi_deep_2018,sitzmann_deepvoxels:_2019,tewari2020state,mildenhall2020nerf}, in which a separate neural network is trained for each subject to be rendered, and all images of that subject are treated as ``training data.''
These approaches have shown great promise in terms of their rendering fidelity, but they require per-subject training and are unable to generalize across subjects yet.
Unlike prior work that focuses on a specific task (\eg, relighting or view synthesis), our texture-space formulation allows for simultaneous light and view interpolation.
Furthermore, our model is a valuable training data generator for many works that rely on high-quality renderings of subjects under arbitrary lighting conditions and from multiple viewpoints, such as \cite{meka_deep_2019,sun_single_2019,simply_lookingood,kim2018deep,BMSengupta20}.
\section{Method}
\label{sec:method}

\begin{figure}[!htbp]
    \vspace{-0.2cm}
\centering
    \includegraphics[width=\linewidth]{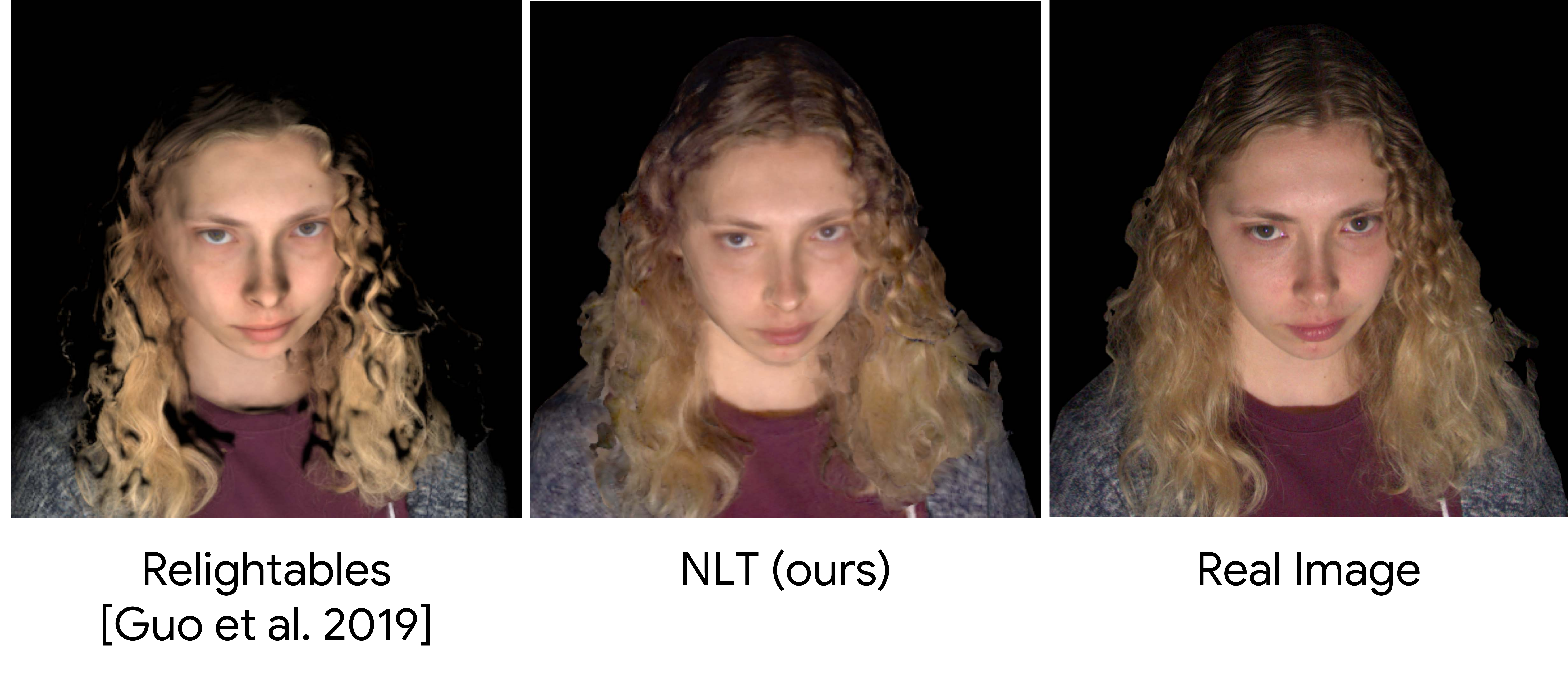}
    \vspace{-0.75cm}
    \caption{\textbf{\rev{Gap in photorealism} between a traditional rendering and a real image.} Even when high-quality geometry and albedo can be captured (\eg, by Relightables~\citep{guo_relightables:_2019}), photorealistic rendering remains challenging, because any geometric inaccuracy will show up as visual artifacts (\eg, black rims/holes in the hair), and manually creating spatially-varying, photorealistic materials is onerous, if possible at all. \model aims to close this gap by learning directly from real images the residuals that account for geometric inaccuracies and non-diffuse LT, such as global illumination.}
  \label{fig:gap}
    \vspace{-0.3cm}
\end{figure}

Our framework is a semi-parametric model with a residual learning scheme that aims to close the gap between the diffuse rendering of the geometry proxy and the real input image (Figure~\ref{fig:gap}). The semi-parametric approach is used to fuse previously recorded observations to synthesize a novel, photorealistic image under any desired illumination and viewpoint.

The method relies on recent advances in computer vision that have enabled accurate 3D reconstructions of human subjects, such as the technique of \citet{fvv} which takes as input several images of a subject and produces as output a mesh of that subject and a UV texture map describing its albedo. At first glance, this appears to address the entirety of our problem: given a textured mesh, we can perform simultaneous view synthesis and relighting by simply re-rendering that mesh from some arbitrary camera location and under some arbitrary illumination. However, this simplistic model of reflectance and illumination only permits equally simplistic relighting and view synthesis, assuming Lambertian reflectance:
\begin{equation}
\tilde{L}_{o}(\mathbf{x}, \viewdir) = \rho(\mathbf{x}) L_i\left(\mathbf{x}, \outdir\right) \left(\outdir \cdot \mathbf{n}(\mathbf{x})\right) .
\label{eq:lambertian}
\end{equation} 
Here $\tilde{L}_{o}(\mathbf{x}, \viewdir)$ is the diffuse rendering of a point $\mathbf{x}$ with a surface normal $\mathbf{n}(\mathbf{x})$ and albedo $\rho(\mathbf{x})$, lit by a directional light $\outdir$ with an incoming intensity $L_i\left(\mathbf{x}, \outdir\right)$ and viewed from $\viewdir$. %
This reflectance model is only sufficient for describing matte surfaces and direct illumination. More recent methods (such as Relightables~\cite{guo_relightables:_2019}) also make strong assumptions about materials by modeling reflectance with a cosine lobe model. The shortcomings of these methods are obvious when compared to a more expressive rendering approach, such as the rendering equation \cite{Kajiya_rendeq_1986}, which makes far fewer simplifying assumptions:
\begin{align}
L_{o}(\mathbf{x}, \viewdir)&=L_{e}(\mathbf{x}, \viewdir)\ + \nonumber\\
&\int_{\Omega} f_{s}\left(\mathbf{x}, \outdir, \viewdir\right) L_i\left(\mathbf{x}, \outdir\right)\left(\outdir \cdot \mathbf{n}(\mathbf{x})\right)\,\operatorname {d} \outdir .
\label{eq:renderingeq}
\end{align}

From this we observe the many limitations in computing $\tilde{L}_{o}(\mathbf{x}, \viewdir)$: it assumes a single directional light instead of integrating over the hemisphere of all incident directions $\Omega$, it approximates an object's BRDF $f_{s}(\cdot)$ as a single scalar, and it ignores emitted radiance $L_{e}(\cdot)$ (in addition to scattering and transmittance, which this rendering equation does not model either). 
The goal of our learning-based model is to close the gap between $L_{o}(\mathbf{x}, \viewdir)$ and $\tilde{L}_{o}(\mathbf{x}, \viewdir)$, and furthermore between $L_{o}(\mathbf{x}, \viewdir)$ and the observed image.

Though not perfect for relighting, the geometry and texture atlas provided by \citet{guo_relightables:_2019} offers us a mapping from each image of a subject onto a canonical texture atlas that is shared across all views of that subject. This motivates the high-level approach of our model: we use this information to map the input images of the subject from ``camera space'' (XY pixel coordinates) to ``texture space'' (UV texture atlas coordinates), then use a semi-parametric neural network \emph{embedded} in this texture space to fuse multiple observations and synthesize an RGB texture atlas for the desired relit and/or novel-view image. This is then warped back into the camera space of the desired viewpoint, thereby giving us an output rendering of the subject under the desired illumination and viewpoint.

In Section~\ref{sec:data} and Section~\ref{sec:uvbuffers}, we describe our data acquisition setup and the input data to our framework. In Section~\ref{sec:queryobs}, we detail the texture-space two-path neural network architecture at the core of our model, which consists of: 1) ``\obspath'' that take as input a set of observed RGB images that have been warped into the texture space and produce a set of intermediate neural activations, and 2) a ``\querypath'' that uses these activations to synthesize a texture-space rendering of the subject according to some desired light and/or viewing direction. 

\begin{figure*}[!htbp]
  \centering
  \includegraphics[width=\textwidth]{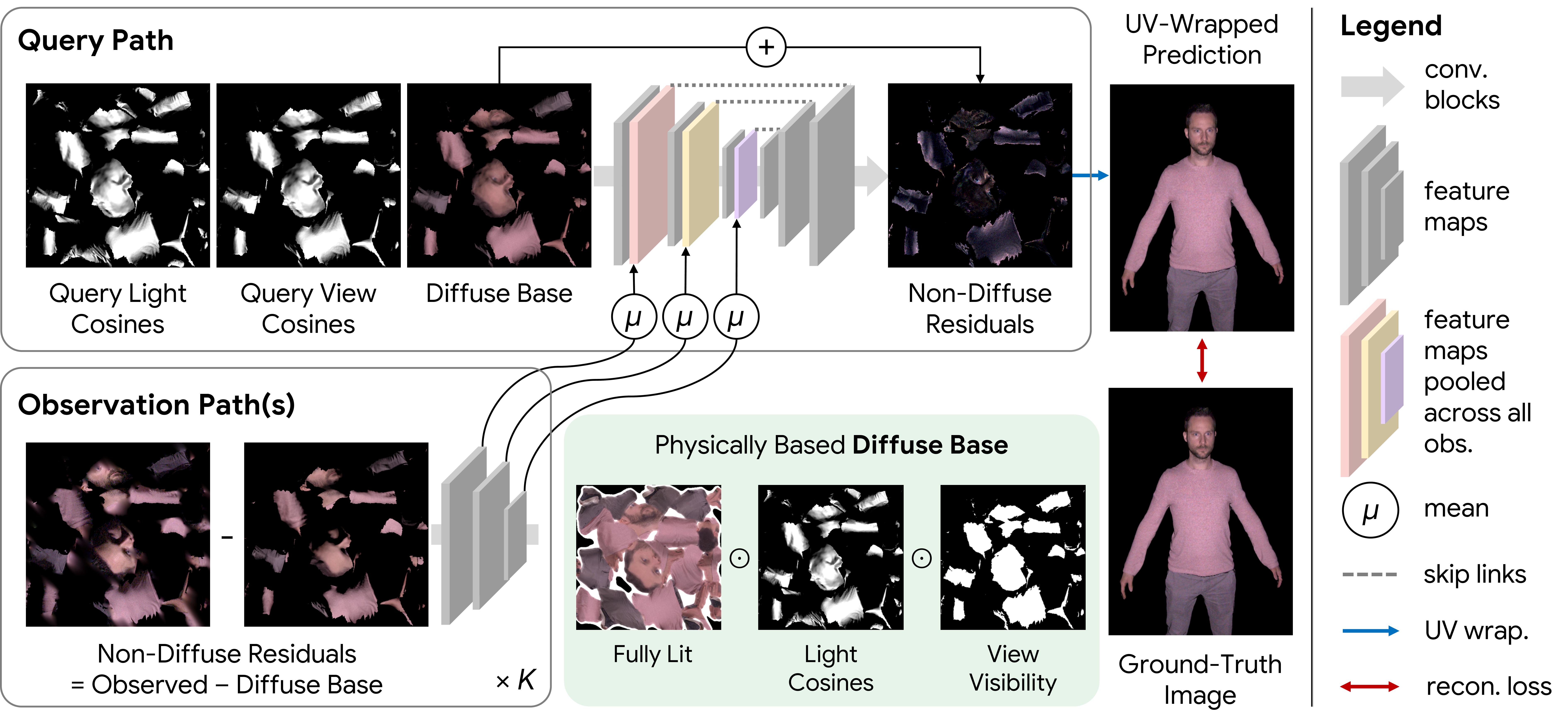}
  \vspace{-0.7cm}
  \caption{
  \textbf{Model.}
  Our network consists of two paths. 
  The ``\obspath'' take as input $K$ nearby observations (as texture-space residual maps) sampled around the target light and viewing directions, and encode them into multiscale features that are pooled to remove the dependence on their order and number.
  These pooled features are then concatenated to the feature activations of the ``\querypath,'' which takes as input the desired light and viewing directions (in the form of cosine maps) as well as the physically accurate diffuse base (also in the texture space). This path predicts a residual map that is added to the diffuse base to produce the texture rendering. With the (differentiable) UV wrapping pre-defined by the geometry proxy, we then resample the texture-space rendering back into the camera space, where the prediction is compared against the ground-truth image. Because the entire network is embedded in the texture space of a subject, the same model can be trained to perform relighting, view synthesis, or both simultaneously, depending on the input and supervision. %
  }
  \label{fig:model}
  
    \vspace{-0.2cm}
\end{figure*}

The texture-space inputs encode a rudimentary geometric understanding of the scene and correspond to the arguments of the 6D LT function (\ie, UV location on the 3D surface $\mathbf{x}$, incident light direction $\outdir$, and viewing direction $\viewdir$).
By using a skip-link between the \querypath's diffuse base image and its output as described in Section~\ref{sec:residual}, our model is encouraged to learn a residual between the provided Lambertian rendering with geometric artifacts and the real-world appearance, which not only guarantees the physical correctness of the diffuse LT, but also directs the network's attention towards learning higher-order, non-diffuse LT effects.
In Section~\ref{sec:loss}, we explain how our model is trained end-to-end to minimize a photometric loss and a perceptual loss in the camera space.
Our model is visualized in Figure~\ref{fig:model}.

\subsection{Hardware Setup and Data Acquisition}
\label{sec:data}

\begin{figure}[!htbp]
\vspace{-0.1cm}
\centering
    \includegraphics[width=\linewidth]{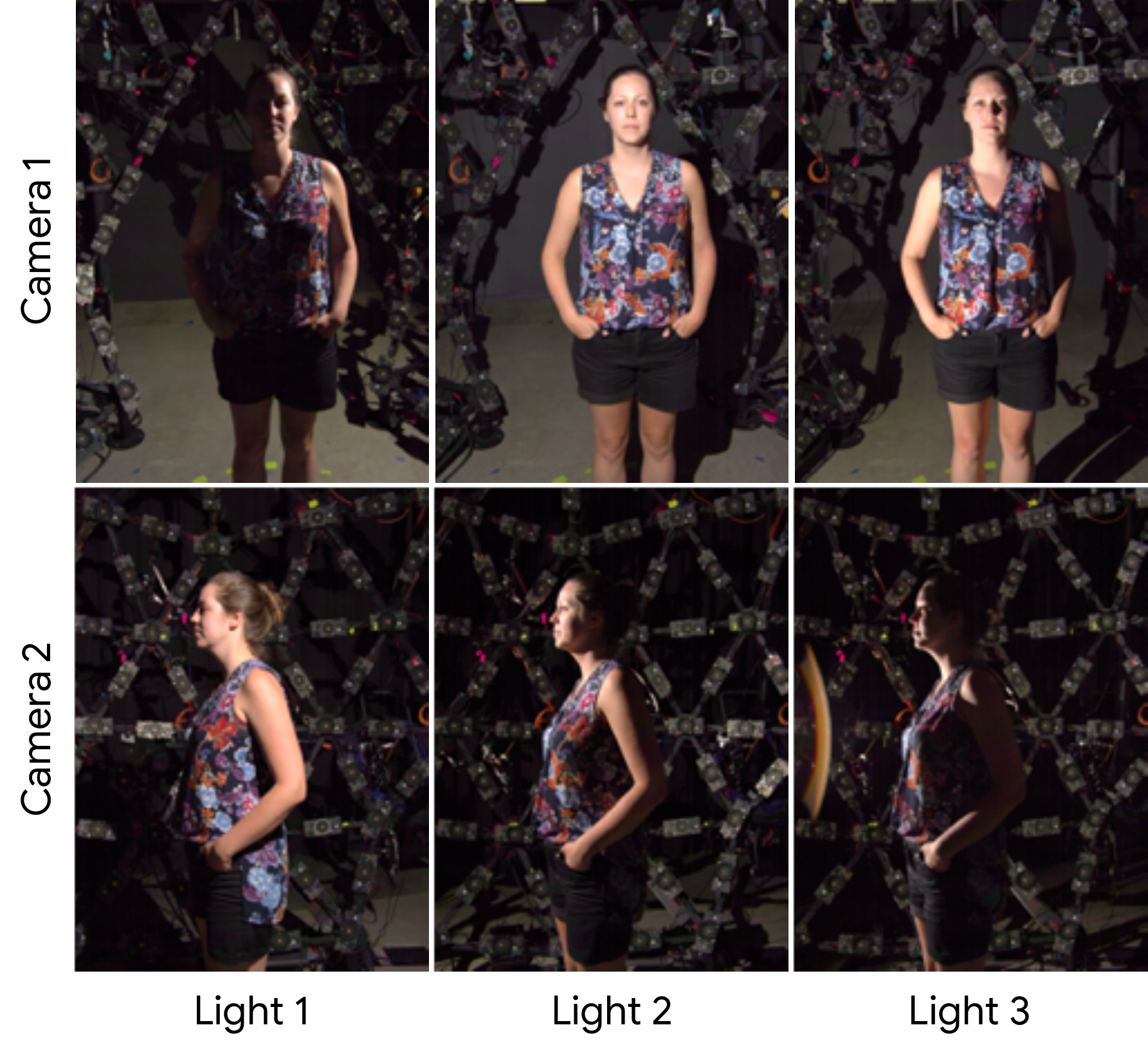}
    \vspace{-0.75cm}
    \caption{\textbf{Sample images used for model training.} We train our model with multi-view, One-Light-at-A-Time (OLAT) images, in each of which only one (known) directional light is active at a time. A \emph{proxy} of the underlying geometry is also required by \model, but it can be as rough as 500 vertices (see Section~\ref{sec:analysis}). These images are sparse samples of the 6D light transport function that \model learns to interpolate.}
  \label{fig:data}
   \vspace{-0.2cm}
\end{figure}

Our method relies on directional light images (One-Light-at-A-Time [OLAT] images) in the form of texture-space UV buffers. 
This requires us to acquire training images under known illumination conditions alongside a parameterized geometric model to obtain the UV buffers.
We use a light stage similar to \citet{sun_single_2019} to acquire OLAT images where only one (known) directional light source is active in each image. For each session we captured $331$ OLAT images for $64$ RGB cameras placed around the performer. When a light is pointing towards a given camera or gets blocked by the subject, the resultant image is either ``polluted'' by the glare or is overly dark. As such, for a given camera, there are approximately 130 usable OLAT images. These OLAT images are sparse samples of the 6D light transport function, which \model learns to interpolate. We visualize samples of these OLAT images in \fig{fig:data}.

Following previous works~\cite{meka_deep_2019,sun_single_2019}, we ask the subject to stay still during the acquisition phase, which lasts $\sim$6 seconds for a full OLAT sequence. Since it is nearly impossible for the performer to stay perfectly still, we align all the images using the optical flow technique of \citet{meka_deep_2019}:
we capture ``all-lights-on'' images throughout the scan that are used as ``tracking frames,'' and compute 2D flow fields between each tracking frame and a reference tracking frame taken from the middle of the sequence. 
These flow fields are then interpolated from the tracking frames to the rest of the images to produce a complete alignment.

Following the approach of \citet{guo_relightables:_2019}, we use $32$ high-resolution active IR cameras and $16$ custom dot illuminator projectors to construct a high-quality parameterized base mesh of each subject fully automatically. These data are critical to our approach, as the estimated geometry provided by this system provides the substrate that our learned model is embedded within in the form of a texture atlas. However, this captured 3D model is far from perfect due to approximations in the mesh model (that cannot accurately model fine structures such as hair) and hand-crafted priors in the reflectance estimation (that relies on a cosine lobe BRDF model). This is demonstrated in~\fig{fig:gap}. Our model overcomes these issues and enables photorealistic renderings, as demonstrated in Section~\ref{sec:experiments}. Additionally, we demonstrate in Section~\ref{sec:analysis} that our neural rendering approach is robust to geometric degradation and can work with geometry proxies of as few as 500 vertices.

We collect a dataset of $70$ human subjects with fixed poses, each of which provides $\sim$\! 18,000 frames under $331$ lighting conditions and $55$ viewpoints (before filtering out glare-polluted and overly dark frames, as aforementioned). We randomly hold out $6$ lighting conditions and $2$ viewpoints for training. Subjects are selected to maximize diversity in terms of clothing, skin color, and age. 
By training our model to reproduce held-out images from these light stage scans, it is able to learn a general LT function that can be used to produce renderings for arbitrary viewpoints and illuminations. Because our scans do not share the same UV parameterization, we train a separate model for each subject.

\subsection{Texture-Space Inputs}
\label{sec:uvbuffers}

In order to perform light and view interpolation, we use as input to our model a set of OLAT images, the subject's diffuse base, and the dot products of the surface normals with the desired or observed viewing directions or light directions (\aka ``cosine maps''), all in the UV texture space.
This augmented input allows our learned model to leverage insights provided by classic graphics models, as the dot products between the normals and the viewing or lighting directions are the standard primitives in parametric reflectance models (Equations~\ref{eq:lambertian}, \ref{eq:renderingeq}, \etc).

These augmented texture-space input buffers superficially resemble the ``G-buffers'' used in deferred shading models~\cite{gbuffer} and used with neural networks in Deep Shading~\cite{Nalbach2017b}.
But unlike \citet{Nalbach2017b}, our goal is to train a model for view and light interpolation using \emph{real images}, instead of renderings from a CG model. This different goal motivates additional novelties of our approach, such as the embedding of our model in UV space (which removes the dependency on viewpoints and implicitly provides aligned correspondence across multiple views) and our use of a residual learning scheme (to encourage training to focus on higher-order LT effects).
\citet{li19} also successfully employ deep learning in the texture space and regress PRT coefficients for deformable objects, but they learn only predefined diffuse and glossy light transport from synthetic rendering.
We use three types of buffers in our model, as described below.

\myparagraph{Cosine map}
Assuming directional light sources, we calculate the cosine map of a light as the dot product between the light's direction $\light$ and the surface's normal vector $\mathbf{n}(\mathbf{x})$.
For each view and light (both observed and queried), we compute two cosine maps: a view cosine map $\mathbf{n}(\mathbf{x}) \cdot \viewdir$ and a light cosine map $\mathbf{n}(\mathbf{x}) \cdot \outdir$.
Crucially, these maps \rev{are masked by visibility computed via ray casting from each camera onto the geometry proxy}, such that the light cosines also provide rough understanding of cast shadows \rev{(texels with zero visibility; see~\fig{fig:model})}, leaving the network an easier task of adding, \eg, global illumination effects to these hard shadows.

\myparagraph{Diffuse base}
The diffuse base is obtained by summing up all OLAT images for each view or equivalently, illuminating the subject from all directions simultaneously (because light is additive). These multiple views are then averaged together
in the texture space, which mitigates the view-dependent effects and produces a texture map that resembles albedo. Note that multiplying the diffuse base by a light cosine map produces the diffuse rendering (with hard cast shadows) for that light $\tilde{L}_{o}(\mathbf{x}, \outdir)$. The construction of this diffuse base is visualized in the bottom middle of Figure~\ref{fig:model}.

\myparagraph{Residual map}
We compute the difference between each observed OLAT image and the aforementioned diffuse base, thereby capturing the ``non-diffuse and non-local'' residual content of each input image. These residual maps are available only for the \textit{sparsely} captured OLAT from fixed viewpoints. To synthesize a novel view for any desired lighting condition, our network uses a semi-parametric approach that interpolates previously seen observations and their residual maps, generating the final rendering.

\subsection{Query and Observation Networks}
\label{sec:queryobs}

Our semi-parametric approach is shown in~\fig{fig:model}: the network takes as input multiple UV buffers in two distinct branches, namely a ``\querypath'' and ``\obspath.'' The \querypath takes as input a set of texture maps that can be generated from the captured geometry, \ie, view/light cosine maps and a diffuse base. The \obspath represent the semi-parametric nature of our framework and have access to non-diffuse residuals of the captured OLAT images. The two branches are merged in an end-to-end fashion to synthesize an unseen lighting condition from any desired viewpoint.

To synthesize a new image of the subject under a desired lighting and viewpoint, we have access to potentially all the OLAT images from multiple viewpoints. The goal of the \obspath is to combine these images and extract meaningful features that are passed to the \querypath to perform the final rendering. However, using all these observations as input is not practical during training due to memory and computational limits. Therefore, for a desired novel view and light condition, we randomly select only $K=1\text{ or }3$ OLAT images from the ``neighborhood'' as observations (the precise meaning of ``neighborhood'' will be clarified in Section~\ref{sec:loss}). The random sampling prevents the network from ``cheating'' by memorizing fixed neighbors-to-query mappings and encourages it to learn that for a given query, different observation selections should lead to the same prediction \rev{(also observed by~\citet{sun2020light})}.

These observed images (in the form of UV-space residual maps as shown in~\fig{fig:model}) are then fed in parallel (\ie, processed as a ``batch'') into the \obspath of our network, which can alternatively be thought of as $K$ distinct networks that all share the same weights. The resulting set of $K$ network activations are then averaged across the set of images by taking their arithmetic mean\footnote{In practice, we observe no improvement when we replace the uniform weights with the  barycentric coordinates of the query \wrt its $K=3$ observations.}, thereby becoming invariant to their cardinality and order, and are then passed to the \querypath.

While the goal of the \obspath is to process input images and glean reflectance information from them, the goal of the \querypath is to take as input information that encodes the non-diffuse residuals of nearby lights/views and then predict radiance values of the queried light and view positions at each UV location.
We therefore concatenate the aggregated activations from the \obspath to the self-produced activations of the \querypath using a set of cross-path skip-connections. The \querypath then decodes a texture-space rendering of the subject under the desired light and viewing directions, which is then resampled to the perspective of the desired viewpoint using conventional, differentiable UV wrapping.

Our proposed architecture has several advantages over a single-path network that would take as input all the available observations, which would be prohibitively expensive in terms of memory and computation.
Because our \obspath do not depend on a fixed order or number of images, during training, we can simply select a dynamic subset of whatever observations that are best suited to the desired lighting and viewpoint.
This ability is useful because the lights and cameras in our dataset are sampled at different rates---lights are $\around 4\times$ denser than cameras. %
The superiority of this dual-path design is demonstrated by both qualitative and quantitative experiments in~\sect{sec:ablation}.

\subsection{Residual Learning of High-Order Effects}
\label{sec:residual}

\begin{figure*}[!htbp]
  \centering
  \includegraphics[width=\textwidth]{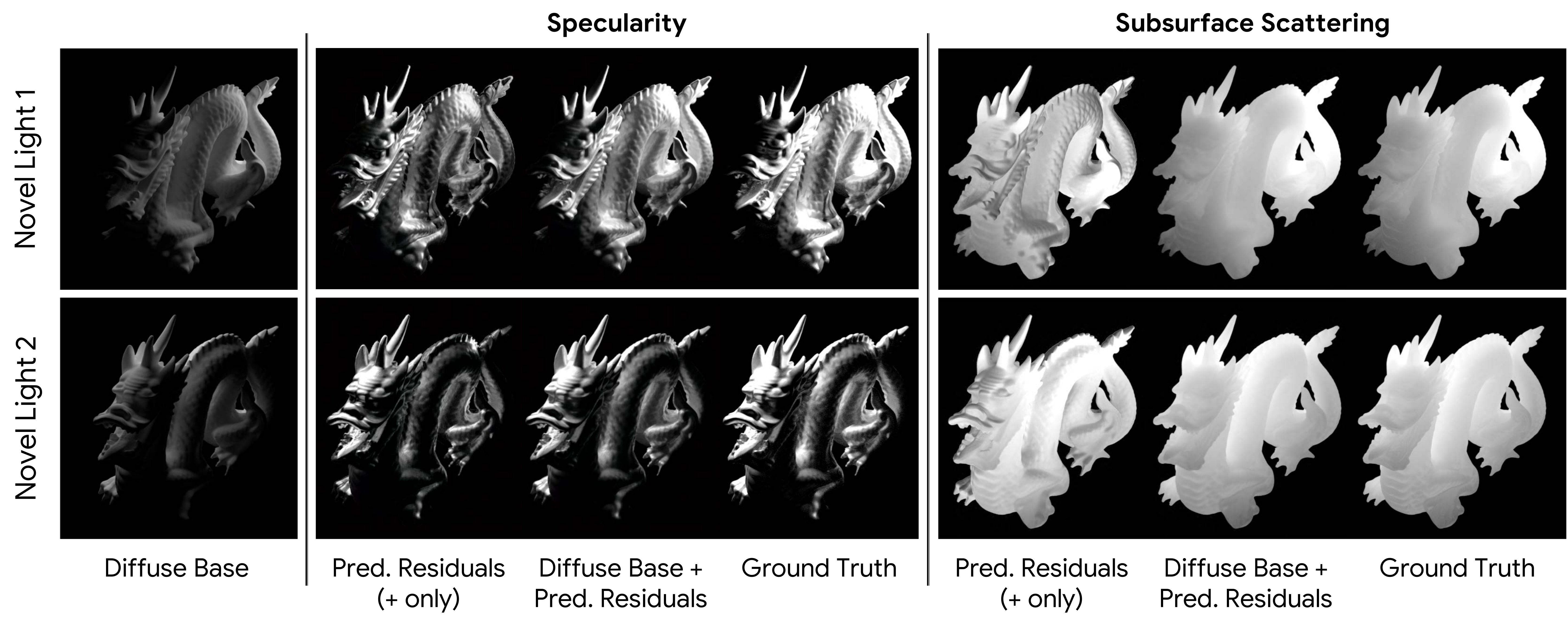}
  \vspace{-0.7cm}
  \caption{
  \textbf{Modeling non-diffuse BSSRDFs as residuals in relighting.}
  A diffuse base (left) captures all diffuse LT (\eg, hard shadows) under a novel point light. 
  By learning a residual on top of this base rendering, \model can reproduce non-diffuse LT (here, specularities and subsurface scattering) from the actual scene appearance.
  When predicting specularities (center), the \model emits exclusively positive residuals (negative part hence not shown) to add bright highlights to the diffuse base.
  When predicting scattering (right), the additive residuals represent additional illumination provided by nearby subsurface light transport.
  }
  \label{fig:lt_demo_specular_sss}
  
   \vspace{-0.1cm}
\end{figure*} 
\begin{figure}[!htbp]
  \centering
  \includegraphics[width=\linewidth]{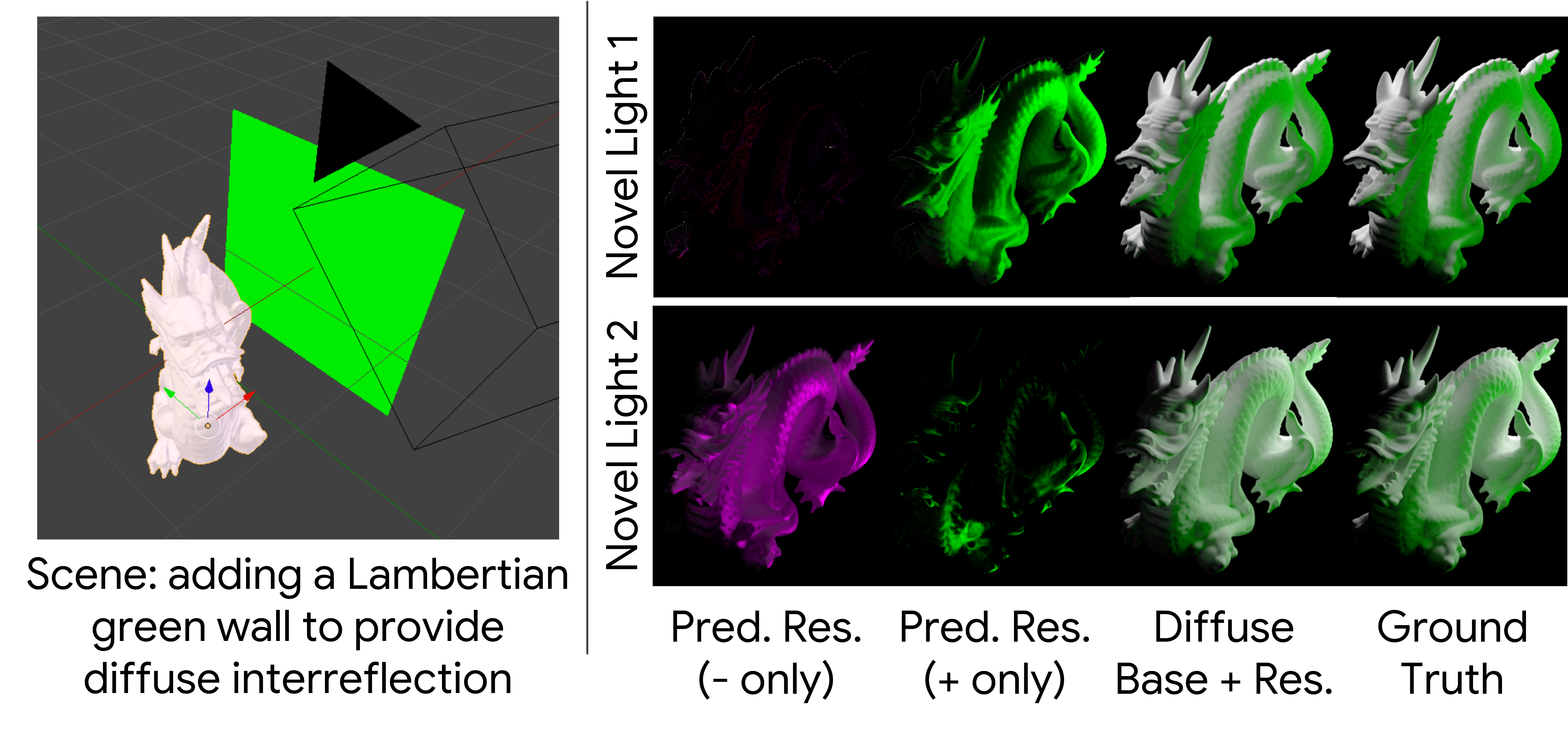}
  \vspace{-0.7cm}
  \caption{
  \textbf{Modeling global illumination as residuals in relighting} (same diffuse bases as in~\fig{fig:lt_demo_specular_sss}). In addition to intrinsic material properties, \model can also learn to express global illumination (\eg, diffuse interreflection) as residuals. Here we add a diffuse green wall to the right of the scene (left). Under Novel Light 1 (right top), the wall provides additional green indirect illumination, so the residuals are green and mostly positive. Notably, the residuals are not necessarily all positive: under Novel Light 2 (right bottom), the residuals are mostly negative and high in blue and red, effectively casting ``negative purple'' indirect illumination that results in a greenish tinge.
  }
  \label{fig:lt_demo_interreflect}
  \vspace{-0.4cm}
\end{figure} 

When synthesizing the output texture-space image in the \querypath of our network, we do not predict the final image directly. Instead, we have a residual skip-link \cite{he2016deep} from the input diffuse base to the output of our network.
Formally, we train our deep neural network to synthesize a residual $\Delta L$ that is then added to our diffuse base $\tilde{L}_{o}(\mathbf{x}, \viewdir)$ to produce our final predicted rendering $L_{o}(\mathbf{x}, \viewdir) = \Delta L + \tilde{L}_{o}(\mathbf{x}, \viewdir)$.
This approach of adding a physically-based diffuse rendering allows our network to focus on learning higher-order, non-diffuse, non-local light transport effects (specularities, scattering, \etc) instead of having to ``re-learn'' the fundamentals of image formation (basic colors, rough locations and shapes of cast shadows, \etc).
Because these residuals are the unconstrained output of a network, this model is able to describe any output image: positive residuals can be added to represent specularities, and negative residuals can be added to represent shading or shadowing.
This residual approach causes our model to be implicitly regularized towards a simplified but physically-plausible diffuse model -- the network can ``fall back'' to the diffuse base rendering by simply emitting zeros.

We demonstrate that our method is capable of modeling complicated lighting effects including specular highlights (BRDFs), subsurface scattering (BSSRDFs), and diffuse interreflection (global illumination), in the context of relighting a toy dragon scene.
We consider a 3D model with perfect geometry and known material properties and render it in a virtual scene similar to a light stage setup using Cycles (Blender's built-in, physically-based renderer).
We produce a diffuse render of the scene as a baseline, and then re-render it using both our model and Blender with three lighting effects: specular highlights, subsurface scattering, and diffuse interreflections, to demonstrate that \model is capable of modeling those effects. The results are shown in \fig{fig:lt_demo_specular_sss} and \fig{fig:lt_demo_interreflect}. %

\myparagraph{Specular highlights}
In Blender, we mix a glossy shader into the dragon's diffuse shader and re-render the scene, resulting in a render with highlights. We then train our model to infer these residuals for relighting. In \fig{fig:lt_demo_specular_sss} (center), we show the \model renderings under two novel light directions (unseen during training) alongside with the ground-truth renderings. The residual image predicted by our model correctly models the specular highlights, and our rendering closely resembles the ground truth.

\myparagraph{Subsurface scattering}
Our model can capture lighting effects that cannot be captured by a BRDF, such as subsurface scattering. We mix a subsurface scattering shader into the dragon's diffuse shader, and then train our model to learn these effects in relighting. As shown in~\fig{fig:lt_demo_specular_sss} (right), the \model results are almost identical to the ground truth.

\myparagraph{Diffuse interreflection}
To demonstrate global illumination, in \fig{fig:lt_demo_interreflect} we place a matte green wall into the scene, and we see that \model is able to accurately predict the non-local light transport of a green glow cast by the wall onto the dragon.

\subsection{Simultaneous Relighting and View Synthesis}
\label{sec:simul}
Embedded in the texture space, \model is a unified framework that can perform relighting, view synthesis, or both simultaneously. The architecture described in Section \ref{sec:queryobs} takes as input the cosine maps that encode the light and viewing directions, as well as a set of observed residual maps from nearby lights and/or views (neighbor selection scheme in Section~\ref{sec:loss}). Since there is no model design specific to relighting or view synthesis, the model is agnostic to which task it is solving other than interpolating the 6D LT function. Therefore, by varying both lights and views in the training data, the model can be trained to render the subject under any desired illumination from any camera position (\aka simultaneously relighting and view synthesis). We demonstrate this capability in Section~\ref{sec:viewsyn} and the supplemental video.

\subsection{Network Architecture, Losses, and Other Details}
\label{sec:loss}
\newcommand{\ipred}{L_{o}(\mathbf{x}_i, \viewdir)}
\newcommand{\igt}{L^*_{o}(\mathbf{x}_i, \viewdir)}

Both paths of our architecture are modifications of the U-Net \cite{ronneberger2015u}, where our \querypath is a complete encoder-decoder architecture (with skip-connections) that decodes the final predicted image, while our observation path is just an encoder.
Following standard conventions, each scale of the network consists of two convolutional layers (except at the very start and end), where downsampling and upsampling are performed using strided (possibly transposed) convolutions, and the channel number of the feature maps is doubled after each downsampling and halved after each upsampling. 
Detailed descriptions of the architectures of these two networks are provided in Tables~\ref{arch:obs} and \ref{arch:query}.
$\conv(d, w\times h, s)$ denotes a two-dimensional convolutional layer (\aka ``$\operatorname{conv2D}$'') with $d$ output channels, a filter size of $(w \times h)$, and a stride of $s$, and is always followed by a leaky ReLU~\cite{maas2013rectifier} activation function. \upconv is the transpose of \conv\,
and is also followed by a leaky ReLU. No normalization is used. Note that the activations from the \obspath are appended to the \querypath before its internal skip connections, meaning that observation activations are effectively skip-connected to the decoder of the query network.

\begin{table}[!htbp]
\caption{The neural network architecture of our ``observation path.''}
    \vspace{-.7em}
\begin{tabular}{ r | l | l l r}
 ID & Operator & \multicolumn{3}{c}{Output Shape} \\
 \hline 
 O1 & \conv(16, $1 \times 1$, 1)  & $H$ & $W$ & $16$ \\ 
 O2 & \conv(16, $3 \times 3$, 2)  & $H/2$ & $W/2$ & $16$ \\ 
 O3 & \conv(16, $3 \times 3$, 1)  & $H/2$ & $W/2$ & $16$ \\ 
 O4 & \conv(32, $3 \times 3$, 2)  & $H/4$ & $W/4$ & $32$ \\ 
 O5 & \conv(32, $3 \times 3$, 1)  & $H/4$ & $W/4$ & $32$ \\ 
 $\dots$ & $\dots$ & $\dots$ \\
 O14 & \conv(1024, $3 \times 3$, 2)  & $H/128$ & $W/128$ & $1024$ \\ 
 O15 & \conv(1024, $3 \times 3$, 1)  & $H/128$ & $W/128$ & $1024$ \\
 O16 & \conv(2048, $3 \times 3$, 2)  & $H/256$ & $W/256$ & $2048$ \\ 
 O17 & \conv(2048, $3 \times 3$, 1)  & $H/256$ & $W/256$ & $2048$ \\
 \end{tabular}
\label{arch:obs}
    \vspace{1em}
\caption{The neural network architecture of our ``query path.'' The layers that reflect skip connections from the activations of our observation path in Table~\ref{arch:obs} are highlighted in blue, and U-Net-like skip-links within the query path are  highlighted in green.
}
    \vspace{-.7em}
\begin{tabular}{ r | l | l l r}
 ID & Operator & \multicolumn{3}{c}{Output Shape} \\ \hline 
 Q1 & \conv(16, $1 \times 1$, 1)     & $H$ & $W$ & $16$ \\ 
 \rowcolor{MyLightBlue} Q2 & \append(\mean(O1))                   & $H$ & $W$ & $32$ \\ 
 Q3 & \conv(16, $3 \times 3$, 2)     & $H/2$ & $W/2$ & $16$ \\ 
 Q4 & \conv(16, $3 \times 3$, 1)     & $H/2$ & $W/2$ & $16$ \\ 
 \rowcolor{MyLightBlue} Q5 & \append(\mean(O3))                   & $H/2$ & $W/2$ & $32$ \\ 
 Q6 & \conv(32, $3 \times 3$, 2)    & $H/4$ & $W/4$ & $32$ \\ 
 Q7 & \conv(32, $3 \times 3$, 1)    & $H/4$ & $W/4$ & $32$ \\ 
 \rowcolor{MyLightBlue} Q8 & \append(\mean(O5))                   & $H/4$ & $W/4$ & $64$ \\ 
 $\dots$ & $\dots$ & $\dots$ \\
 \rowcolor{MyLightGreen} Q44 & \append(Q8)                         & $H/4$ & $W/4$ & $80$ \\ 
 Q45 & \upconv(8, $3 \times 3$, 2)  & $H/2$ & $W/2$ & $8$ \\ 
 Q46 & \upconv(8, $3 \times 3$, 1)  & $H/2$ & $W/2$ & $8$ \\ 
 \rowcolor{MyLightGreen} Q47 & \append(Q5)                         & $H/2$ & $W/2$ & $40$ \\ 
 Q48 & \upconv(4, $3 \times 3$, 2)  & $H$ & $W$ & $4$ \\ 
 Q49 & \upconv(4, $3 \times 3$, 1)  & $H$ & $W$ & $4$ \\ 
 \rowcolor{MyLightGreen} Q50 & \append(Q2)                         & $H$ & $W$ & $36$ \\ 
 Q51 & \upconv(3, $1 \times 1$, 1)  & $H$ & $W$ & $3$ \\
\end{tabular}
\label{arch:query}

\vspace{-0.7em}

\end{table} 

We trained our model to minimize losses in the image space between the predicted image $\ipred$ and the ground-truth captured image. To this end, we first resample the UV-space prediction back to the camera space, and then compute the total loss as a combination of a robust photometric loss \cite{barron2019general} and a perceptual loss (LPIPS) \cite{zhang_unreasonable_2018}. 

We use the loss function of \citet{barron2019general} with $\alpha=1$ (also known as pseudo-Huber loss) applied to a CDF9/7 wavelet decomposition \cite{wavelet} in the YUV color space:
\begin{equation}
    \losfun = \sum_i \sqrt{\left(\frac{\operatorname{CDF}\left(\operatorname{YUV}\left(\ipred - \igt \right)\right)}{c}\right)^2 + 1} - 1.
\end{equation}
We empirically set the scale hyperparameter $c = 0.01$. As was demonstrated in \citet{barron2019general}, we found that imposing a robust loss in this YUV wavelet domain produced reconstructions that better captured both high-frequency and low-frequency details.

The perceptual loss $\losfunp$ \cite{zhang_unreasonable_2018} is defined as the $\ell_2$ distance in feature space extracted with a VGG network~\cite{simonyan2014very} pre-trained on ImageNet~\cite{imagenet_cvpr09}. The final loss function is simply the sum of the two losses $\mathcal{L}= \losfun + \losfunp$. Empirically, we found that using the same weight for both losses achieved the best results.

We trained our model by minimizing $\mathcal{L}$ using Adam~\cite{KingmaB15} with a learning rate of $2.5\times 10^{-4}$, a batch size of $1$, and the following optimizer hyperparameters: $\beta_1=0.9, \beta_2=0.999, \epsilon=10^{-7}$.
Our model is implemented in TensorFlow~\cite{tensorflow} and trained on a single Nvidia Tesla P100, which takes less than 12 hours for the real scenes and much less for synthetic scenes.

\myparagraph{Observations}
Our \obspath are designed to be invariant to the number and order of observations, so that one can input as many observations into the path as the task requires or memory constraints permit.
During training, the observations are the $K=3$ neighbors randomly sampled from a cone around the query light direction with an apex angle of $30\degree$ for relighting, or from a pool of nearby cameras that have large visibility overlap in the UV space (per-texel ``and'' operation) with the query camera for view synthesis.
In training simultaneous models, we use $K=1$ neighbor with both the nearest camera and light as the observation. At test time, thanks to the framework's invariance to the number and order of observations,  we use a fixed set of observations to reduce flickering caused by frequent neighborhood switching, and \rev{observe no visual difference with different sets of observations}. One may also use all of the physical lights and cameras as observations at test time.

\myparagraph{Resolutions}
For relighting and view synthesis, our texture-space images have a resolution of $1024 \times 1024$, and the camera-space images have a resolution of $1536 \times 1128$. For simultaneous relighting and view synthesis, the resolutions used are $512\times 512$ in the UV space and $1024\times 752$ in the camera space.
\section{Results}
\label{sec:experiments}

We perform multiple quantitative and qualitative experiments to demonstrate the efficacy of our model for relighting (Section~\ref{sec:relight}) and view synthesis (Section~\ref{sec:viewsyn}), which are further clarified through performance analysis (Section~\ref{sec:analysis}) and ablation studies (Section~\ref{sec:ablation}). Additionally, we present qualitative results for HDRI relighting and for simultaneous view synthesis and relighting. All results are on unseen target light and viewing directions unless stated otherwise.

\subsection{Relighting}
\label{sec:relight}

We start our evaluation demonstrating that the method can synthesize new lighting conditions of the subject.
Once we have learned the \nltfull of a given scene, we can relight the scene with a novel incident light direction $\outdir$ by simply querying our model.

\begin{table}[!htbp]
\centering
    \caption{\textbf{Relighting.}
    \Model (or its variant) outperforms all baselines in terms of PSNR and LPIPS~\cite{zhang_unreasonable_2018}.
    Although barycentric blending achieves a similar SSIM score~\cite{zhou_wang_image_2004}, it produces inaccurate renderings (such as ghosting shadows) upon visual inspection (Column 4 of Figure~\ref{fig:relight_dir}).
    Ablating the perceptual loss slightly increases PSNR, but degrades rendering quality by losing high-frequency details (such as the facial specularities shown in~\fig{fig:relight_dir}). The numbers are means and 95\% confidence intervals.
    }
    \label{tbl:relight_dir}
    \vspace{-0.7em}
    \resizebox{\linewidth}{!}{ 
    \begin{tabular}{l|ccc}
        \toprule
        Method & PSNR $\uparrow$ & SSIM $\uparrow$ & LPIPS $\downarrow$\\
        \midrule
        Diffuse Base & $30.21\pm .08$ & $.878\pm .003$ & $.102\pm .003$ \\
        Barycentric Blending & $34.28\pm .20$ & $\mathbf{.942\pm .002}$ & $.051\pm .002$ \\
        Deep Shading [2017] & $33.67\pm .27$ & $.918\pm .009$ & $.106\pm .012$ \\
        \citet{xu_deep_2018} & $31.94\pm .09$ & $.923\pm .003$ & $.089\pm .003$ \\
        Relightables [2019] & $31.03\pm .08$ & $.891\pm .003$ & $.090\pm .003$ \\
        \nlt (ours) & $33.99\pm .19$ & $\mathbf{.942\pm .002}$ & $\mathbf{.045\pm .002}$\\
        \midrule[.1pt]
        \nlt w/o residuals & $33.65\pm .25$ & $.928\pm .006$ & $.063\pm .007$ \\
        \nlt w/o observations & $32.56\pm .15$ & $.925\pm .002$ & $.064\pm .002$ \\
        \nlt w/o LPIPS [2018] & $\mathbf{34.43\pm .24}$ & $.939\pm .005$ & $.066\pm .006$ \\
        \bottomrule
    \end{tabular}
    }
\end{table}

\begin{figure*}[!htbp]
  \centering
  \includegraphics[width=\textwidth]{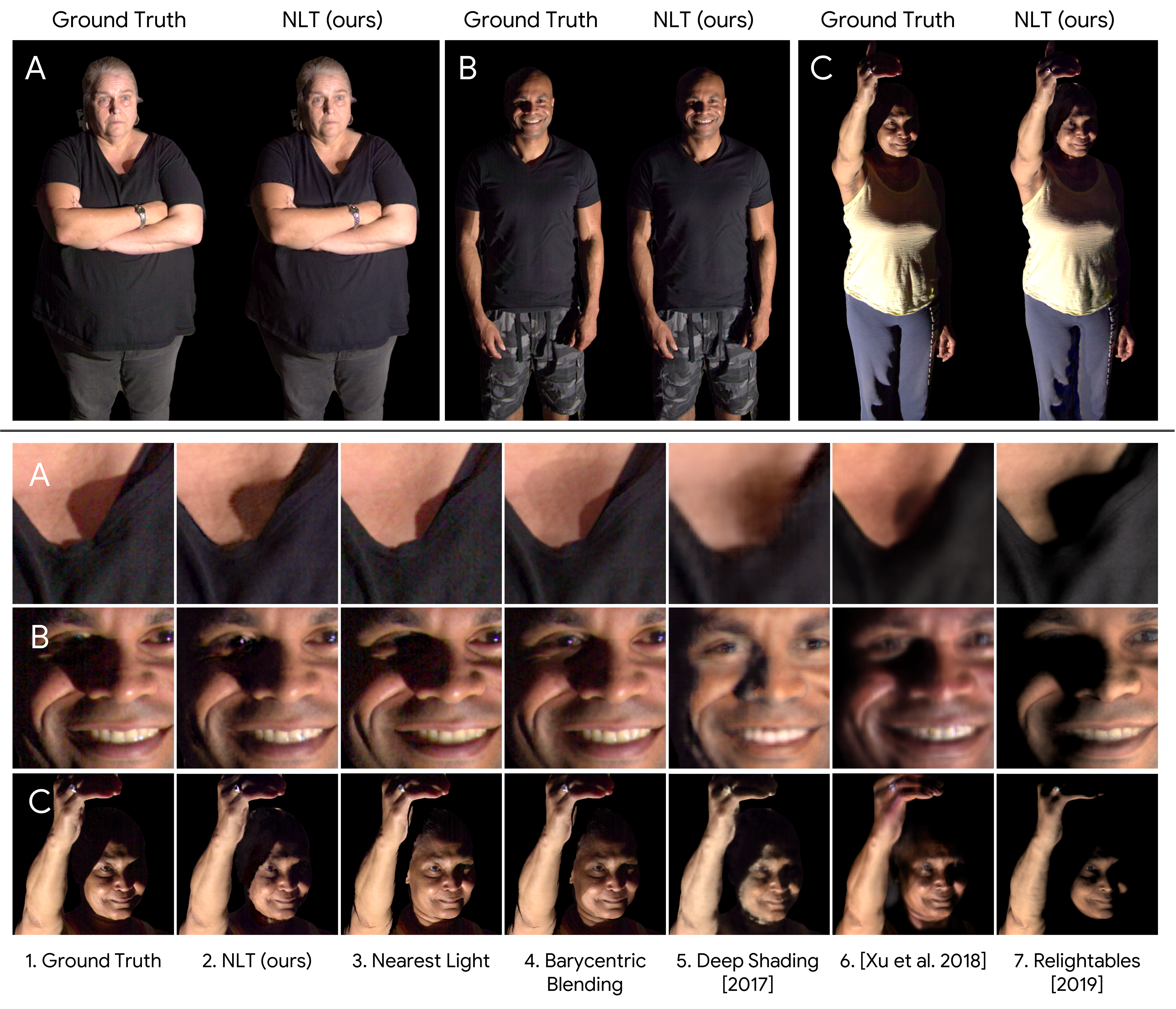}
  \vspace{-0.7cm}
  \caption{\textbf{Relighting by a directional light.}
  Here we visualize the performance of our \model for the task of relighting using directional lights. We show representative examples of full-body subjects with zoom-ins detail focusing on cast shadows (A, C) and facial specular highlights (B). Note how our method is able to outperform all the other approaches with sharper and ghosting-free results that are drastically different from the nearest neighbors.
}
  \label{fig:relight_dir}
  \vspace{-0.1cm}
\end{figure*}

First, we quantitatively evaluate our model against the state-of-the-art relighting solutions and ablations of our model, and report our results in Table~\ref{tbl:relight_dir} in terms of Peak Signal-to-Noise Ratio (PSNR), SSIM~\cite{zhou_wang_image_2004}, and LPIPS~\cite{zhang_unreasonable_2018}.
We see that our method outperforms all baselines and ablations, although simple baselines such as diffuse rendering and barycentric blending also obtain high scores. This appears to be due to these metrics under-emphasizing high frequency details and high-order light transport effects. These results are more easily interpreted using the visualization in Figure~\ref{fig:relight_dir}, where we see that the renderings produced by our approach more closely resemble the ground truth than those of other models. In particular, our method synthesizes shadows, specular highlights, and self-occlusions with higher precision when compared against simple barycentric blending, as well as state-of-art neural rendering algorithms such as \citet{Nalbach2017b} and \citet{xu_deep_2018}. Our approach also produces more realistic results than the geometric 3D capture pipeline of \citet{guo_relightables:_2019}. See the supplemental video for more examples.

\begin{figure*}[!htbp]
  \centering
  \includegraphics[width=\textwidth]{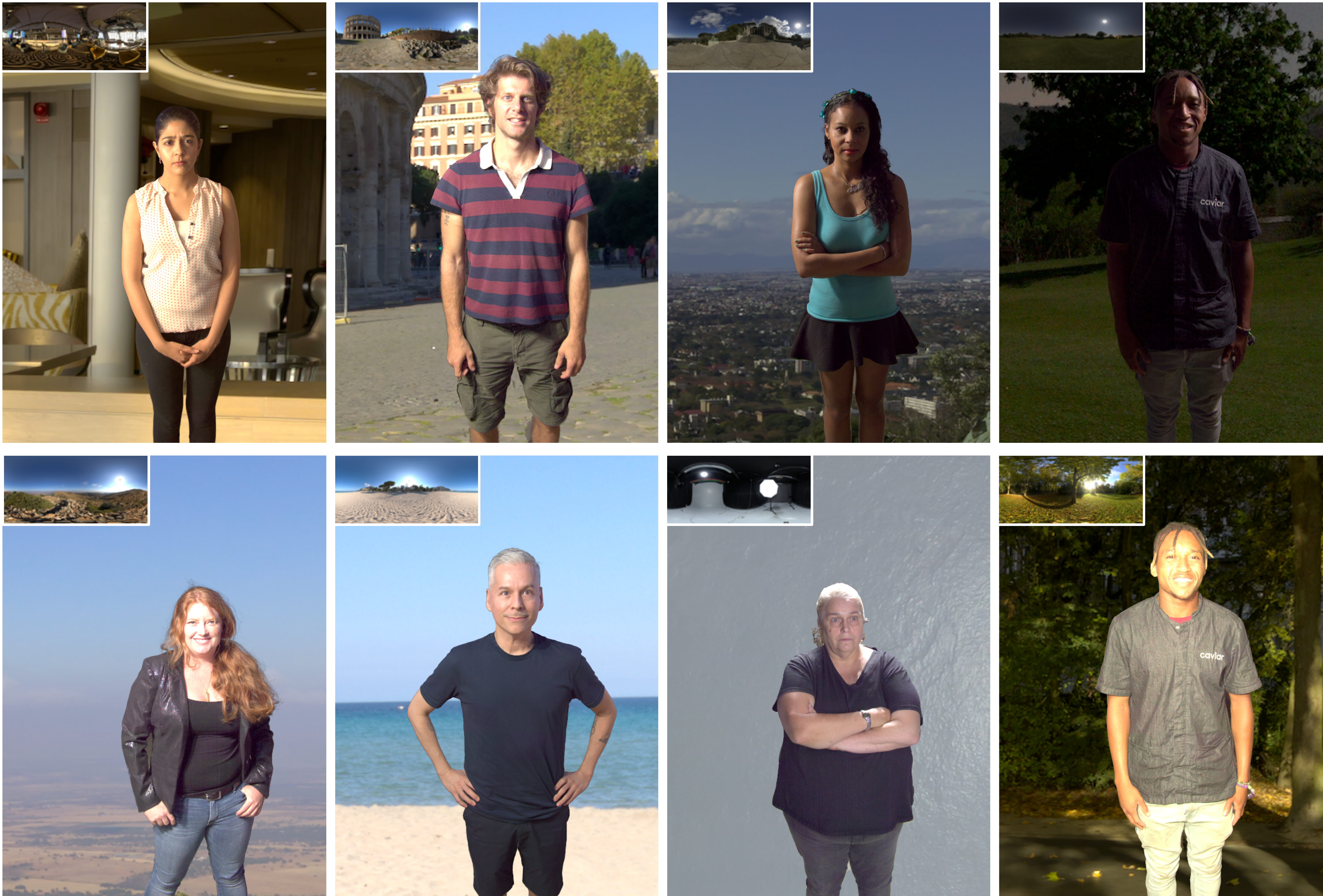}
  \caption{\textbf{HDRI relighting.} Because \model can relight a subject with any directional light, it can be used to render OLAT ``bases'' that can then be linearly combined to relight the scene for a given HDRI map (shown as insets)~\cite{debevec_acquiring_2000}. The relit subjects exhibit realistic specularities and shadows.}
  \label{fig:relight_hdri}
  \vspace{-0.2cm}
\end{figure*} 

\myparagraph{HDRI relighting}
Our model's ability to synthesize images corresponding to arbitrary light directions allows us to render subjects under arbitrary HDRI environment maps.
To do this, we synthesize $331$ directional OLAT images that cover the whole light stage dome. These images are then converted to light stage weights by approximating each light with a Gaussian around its center, and we produce the HDRI relighting results by simply using a linear combination of the rendered OLAT images~\cite{debevec_acquiring_2000}. As shown in \fig{fig:relight_hdri}, we are able to reproduce view-dependent effects as well as specular highlights with high fidelity, and generate compelling composites of the subjects in virtual scenes. See the supplemental video for more examples.

\begin{table}[!htbp]
    \centering
    \caption{\textbf{View synthesis.}
    \Model outperforms all baselines (top) and ablations (bottom) in terms of the LPIPS perceptual metric~\cite{zhang_unreasonable_2018}.
    \rev{UV-space barycentric blending and DNR~\cite{thies_deferred_2019} are competitive in terms of PSNR and SSIM~\cite{zhou_wang_image_2004}, but barycentric blending is unable to hallucinate missing pixels that are not observed from any given view (in contrast to our learning-based approach), and DNR does not provide a means for relighting (simultaneous or otherwise).}
    For fair comparisons, we circumvent the need to learn viewpoints for screen-space methods, such as barycentric blending and Deep Shading~\cite{Nalbach2017b}, by performing their operations in the UV space. The numbers are means and 95\% confidence intervals.}
    \label{tbl:viewsyn}
    \vspace{-0.7em}
    \begin{tabular}{l|ccc}
        \toprule
        Method & PSNR $\uparrow$ & SSIM $\uparrow$ & LPIPS $\downarrow$\\
        \midrule
        Diffuse Base & $31.45\pm .268$ & $.917\pm .005$ & $.070\pm .004$ \\
        Bary.\ Blending (UV) & $34.97\pm .273$ & $.960\pm .002$ & $.035\pm .002$ \\
        Deep Shading (UV) & $34.77\pm .405$ & .$950\pm .009$ & $.058\pm .009$ \\
        DNR & $\mathbf{35.49\pm .315}$ & $\mathbf{.966\pm .002}$ & $.039\pm .002$ \\
        Relightables & $32.24\pm .246$ & $.922\pm .005$ & $.065\pm .004$ \\
        \nlt (ours) & $34.83\pm.259$ & $.959\pm .002$ & $\mathbf{.030\pm .001}$ \\
        \midrule[.1pt]
        \nlt w/o residuals & $34.49\pm .258$ & $.958\pm .002$ & $.032\pm .001$ \\
        \nlt w/o obs.\ & $34.43\pm .286$ & $.953\pm .003$ & $.037\pm .002$ \\
        \nlt w/o LPIPS & $34.36\pm .273$ & $.959\pm .003$ & $.041\pm .002$ \\
        \bottomrule
    \end{tabular}
\end{table} 

\subsection{View Synthesis}
\label{sec:viewsyn}

\begin{figure*}[!htbp]
  \centering
  \includegraphics[width=0.89\textwidth]{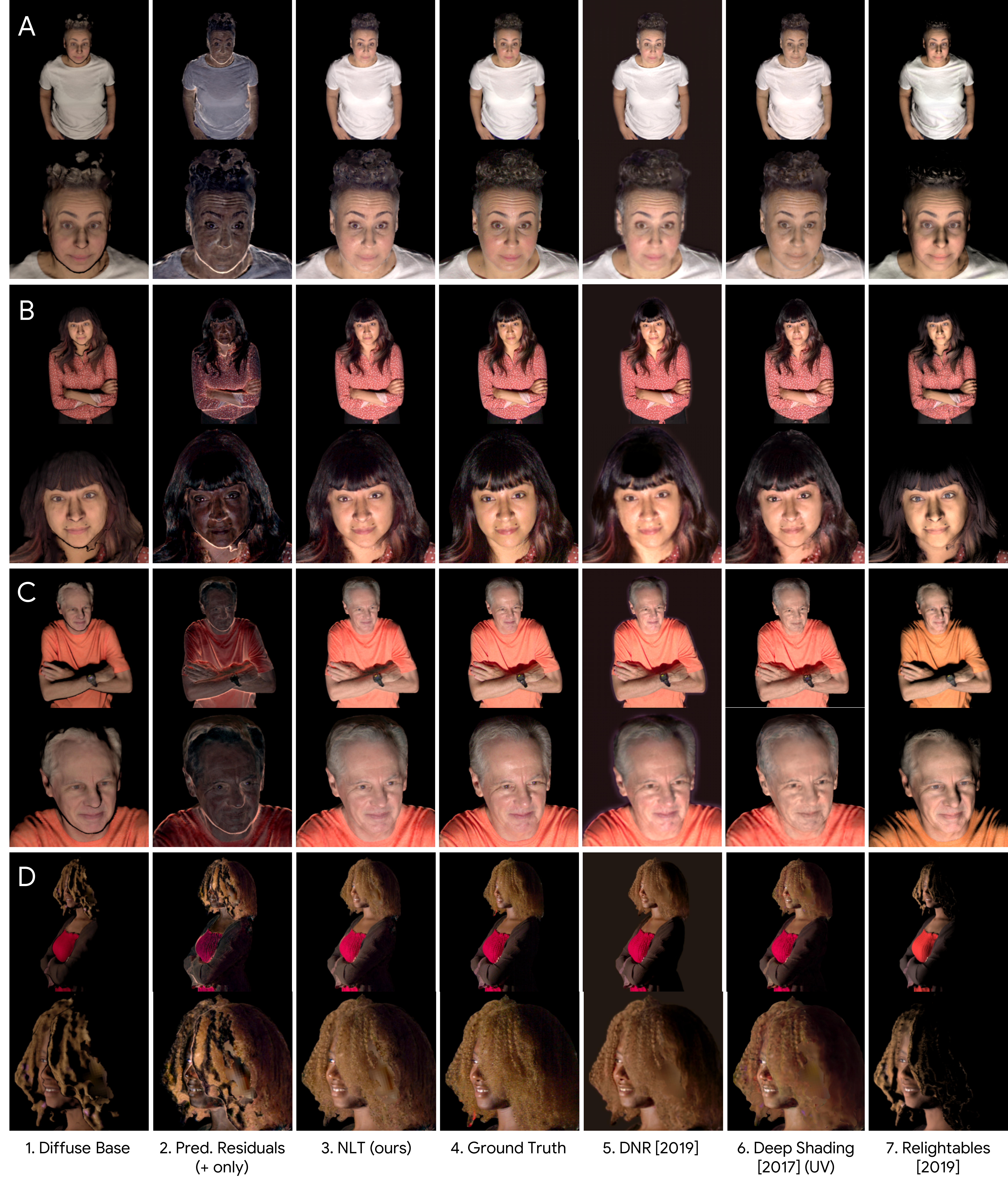}
  \vspace{-0.3cm}
  \caption{\textbf{View synthesis.}
  Here we visualize the  \model results for the task of synthesizing unseen views of a subject.
  The diffuse base (Column 1) fails to capture fine geometry (hair, chins, \etc), non-Lambertian material effects (specularities and subsurface scattering), and global illumination, all of which are corrected for by the residuals (Column 2) predicted by \model (Column 3).
  \model is able to handle view-dependent specularities (eyes, nose tips, cheeks), high-frequency geometry variation (Subjects B's and D's hair), and global illumination (Subjects A, B, and C's shirts).
  We see a substantial improvement over the state-of-the-art view synthesis method of \citet{thies_deferred_2019} (Column 5), which tends to produce blurry results (the missing specularities in Subject B's eyes) %
  and over the recent geometric approach of \citet{guo_relightables:_2019} (Column 7), which lacks non-Lambertian material effects.
  We also compare against an enhanced version of Deep Shading~\cite{Nalbach2017b} that has been trained in our texture space (\`a la~\citet{li19}) so that the model does not need to learn cross-view correspondences.
  As Column 6 shows, images synthesized by this enhanced baseline are less faithful to the ground truth (Column 4) than \model.
  }
  \label{fig:viewsyn}
\end{figure*}

Here we present results for the view synthesis task: synthesizing novel viewpoints while also capturing view-dependent effects.
A quantitative analysis is presented in Table~\ref{tbl:viewsyn}, where we see that our approach outperforms the baselines and is comparable with \citet{thies_deferred_2019}, which (unlike our technique) only performs view synthesis and does not enable relighting.
A qualitative analysis is visualized in Figure~\ref{fig:viewsyn}.
We see that the inferred residuals produced by \model are able to account for the non-diffuse, non-local light transport and mitigate the majority of artifacts in the diffuse base caused by geometric inaccuracy.
We see that renderings from \model exhibit accurate specularities and sharper details, especially when compared with other machine learning methods, thereby demonstrating that our model is able to capture view-dependent effects. See the supplementary video for more examples.

\begin{figure*}[!htbp]
  \centering
  \includegraphics[trim=0 40 0 0, clip, width=\linewidth]{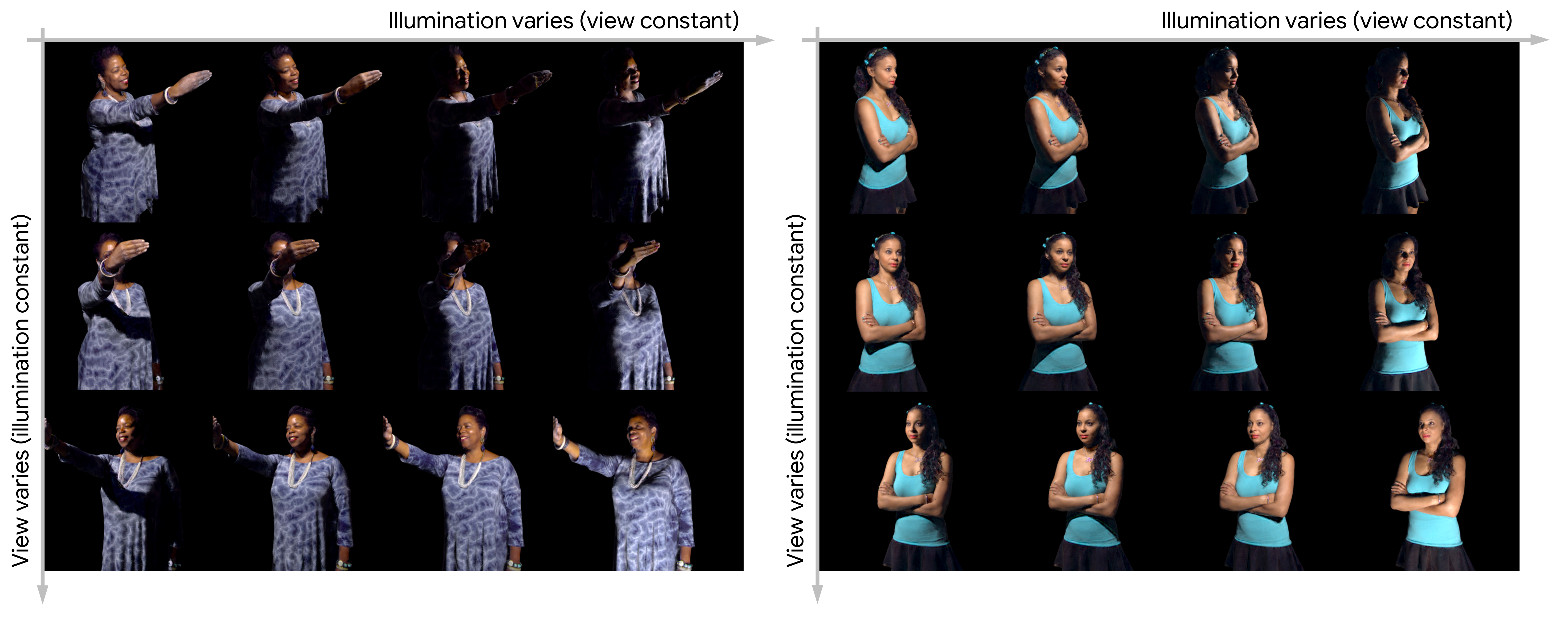}
  \vspace{-0.6cm}
  \caption{\textbf{Simultaneous relighting and view synthesis.}
  Our model is able to perform simultaneous relighting and view synthesis, and produces accurate renderings (including view- and light-dependent effects) for unobserved viewpoints and light directions. In the $x$-axis we vary illumination, and in the $y$-axis we vary the view. 
  This functionality is enabled by our decision to embed our neural network architecture within the texture atlas of a subject.%
  }
  \label{fig:simul}
\end{figure*}

\myparagraph{Simultaneous relighting and view synthesis}
In Figure~\ref{fig:simul}, we show the unique ability of our model to synthesize illumination and viewpoints simultaneously with an unprecedented quality for human capture. Note that our model's ability to naturally handle this simultaneous task is a direct consequence of embedding our neural network within the UV texture atlas space of the subject.
All that is required to enable simultaneous relighting and view interpolation is interleaving the training data for both tasks and training a single instance of our network (more details in Section~\ref{sec:simul}). Figure~\ref{fig:simul} shows that our method accurately models shadows and global illumination, while correctly capturing high-frequency details such as specular highlights. See the supplementary video for more examples.

\begin{figure*}[!htbp]
  \centering
  \includegraphics[width=\linewidth]{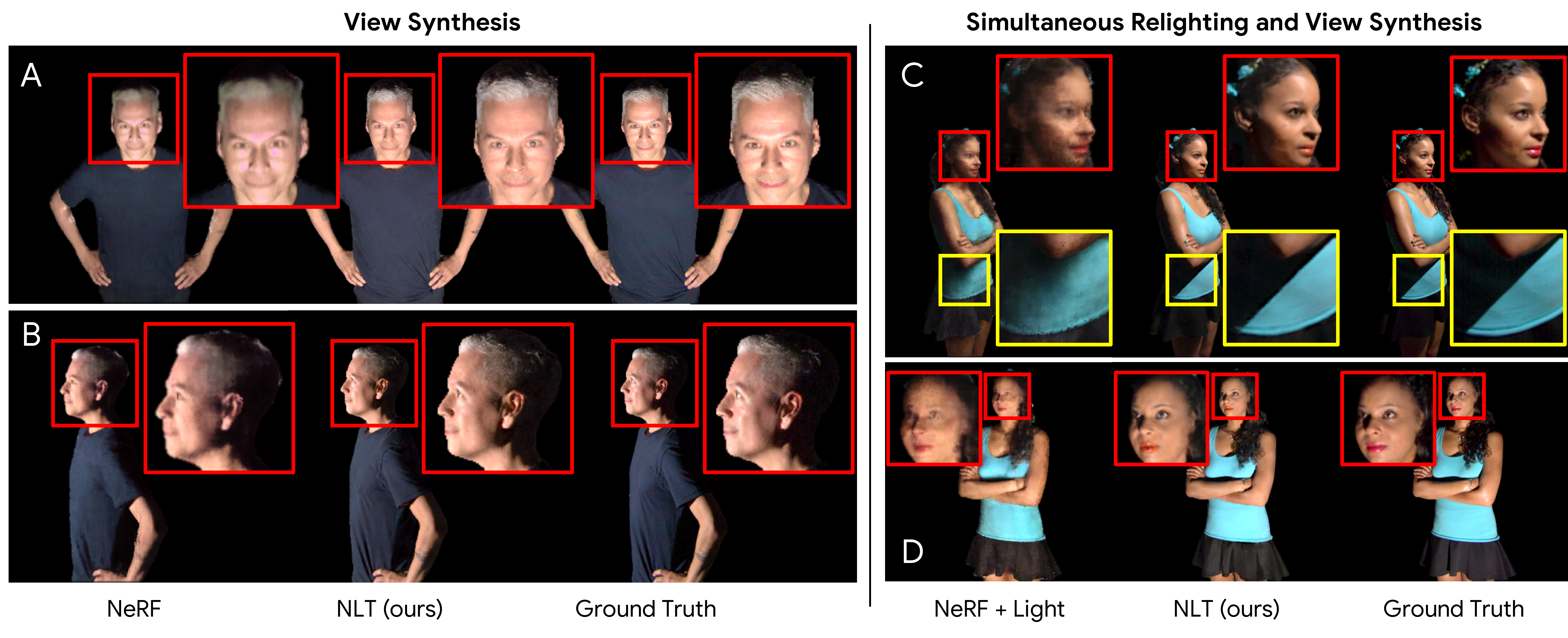}
  \vspace{-0.6cm}
  \caption{
  \rev{
  \textbf{Comparisons with NeRF and a NeRF-based extension that supports simultaneous relighting and view synthesis.}
  In view synthesis (left), NeRF struggles to synthesize realistic facial specularity, high-frequency hair details, and specularity in the eyes (red boxes in A \& B).
  In simultaneous relighting and view synthesis (right), the ``NeRF + Light'' extension fails to synthesize facial details (red boxes in C \& D) and hard shadows (yellow boxes in C).
  }
  }
  \label{fig:nerf}
  \vspace{-0.2cm}
\end{figure*}

\rev{
The recent work of~\citet{mildenhall2020nerf}, Neural Radiance Fields (NeRF), achieves impressive view synthesis given approximately $100$ views of an object. Here we qualitatively compare \model against NeRF with $10$ levels of positional encoding for the location and $4$ for the viewing direction. NeRF does not require any proxy geometry, but in this particular setting, it has to work with a limited number of views (around $55$), which are insufficient to capture the full volume. As~\fig{fig:nerf} (left) shows, \model synthesizes more realistic facial and eye specularity as well as higher-frequency hair details.
}

\rev{
We also attempt to extend NeRF to perform simultaneous relighting and view synthesis, and compare \model with this extension, ``NeRF + Light.'' To this end, we additionally condition the output radiance on the light direction (with $4$ levels of positional encoding) along with the original viewing direction. As shown in~\fig{fig:nerf} (right), NeRF + Light struggles to synthesize hard shadows or specular highlights, and produces significantly more blurry results than \model, which demonstrates the importance of a proxy geometry when there is lighting variation and only sparse viewpoints.
}

\begin{figure*}[!htbp]
\centering
    \includegraphics[width=\linewidth]{fig/perf_viewsyn.pdf}
    \vspace{-0.6cm}
    \caption{
    \textbf{View synthesis performance \wrt quality of the geometry proxy.}
    As we decimate the geometry proxy from 100k vertices down to only 500 vertices, \nlt remains performant in terms of the LPIPS perceptual metric~\cite{zhang_unreasonable_2018} (lower is better; bands indicate 95\% confidence intervals), while Floating Textures~\cite{eisemann2008floating}, a reprojection-based method, suffers from the low quality of the geometry proxy, producing missing pixels (\eg, in the hair) and misplaced high-frequency patterns (\eg, shadow boundaries), as highlighted by the yellow arrows. Both \model and Floating Textures use the same three nearby views.}
  \label{fig:perf_viewsyn}
\end{figure*}

\subsection{Performance Analysis}
\label{sec:analysis}

Here we analyze how our model performs with respect to different factors.
We show that as the geometry degrades, our neural rendering approach consistently outperforms traditional reprojection-based methods, which heavily rely on the geometry quality. In relighting, we show that our model performs reasonably when the number of illuminants is reduced, demonstrating the potential applicability of \model to smaller light stages.

\myparagraph{View synthesis}%
Because \model leverages a geometry proxy to generate a texture parameterization, we study its robustness against geometry degradation in the context of view synthesis. We decimate our mesh progressively from the original 100k vertices down to only 500 vertices (\fig{fig:perf_viewsyn} bottom left). At each mesh resolution, we train one \model model with $K=3$ nearby views and evaluate it on the held-out views. With the geometry proxy, one can also reproject nearby observed views to the query view, followed by different types of blending~\cite{eisemann2008floating,buehler2001unstructured}. We compare renderings from \model with those of \citet{eisemann2008floating} at each decimation level.
As~\fig{fig:perf_viewsyn} shows, even at the extreme decimation level of 500 vertices, \model produces reasonable rendering with no missing pixels, because it has learned to hallucinate pixels that are non-visible from any of the nearby views. In contrast, Floating Textures~\cite{eisemann2008floating} leaves missing pixels unfilled (\eg, in the hair) due to reprojection errors stemming from the rough geometry proxy. As the geometry proxy gets more accurate, Floating Textures improves but still struggles to render high-frequency patterns correctly (such as the shadow boundary beside the nose, highlighted by a yellow arrow), even at the original mesh resolution. In comparison, the high-frequency patterns in the \model rendering match closely the ground truth. Quantitatively, \model also outperforms Floating Textures in terms of the LPIPS perceptual metric ~\cite{zhang_unreasonable_2018} (lower is better) across all mesh resolutions.

\begin{figure}[!htbp]
\centering
    \includegraphics[width=\linewidth]{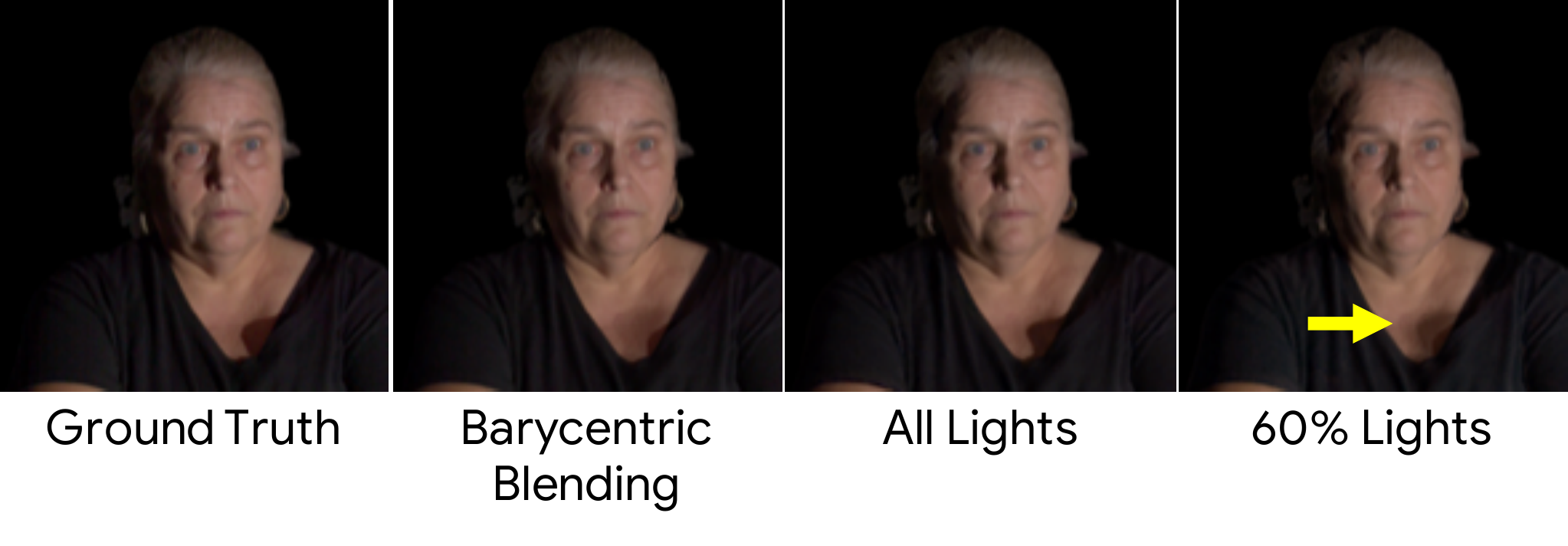}
    \vspace{-0.8cm}
    \caption{\textbf{Training with sparser lights.}
    When only 60\% lights are used to train a relighting model, we observe ghosting shadows in our relit rendering (yellow arrow), similar to those produced by barycentric blending.
    }
  \label{fig:perf_relight}
\end{figure}

\myparagraph{Relighting}%
In this experiment, we artificially downsample the lights on the light stage to study the effects of light density on \model's relighting performance. We use only 60\% of the lights to train a relighting model, which translates to around 75 lights per camera. Although the model is still able to relight the person reasonably in \fig{fig:perf_relight}, inspection reveals that the relit image has ghosting shadows like those often observed in barycentric blending. %

\subsection{Ablation Studies}
\label{sec:ablation}

Our quantitative evaluations of relighting and view synthesis (Tables~\ref{tbl:relight_dir} and \ref{tbl:viewsyn})
include ablation studies of our model, in which separate model components are removed to demonstrate that component's contribution.

\begin{figure}[!htbp]
  \centering
  \includegraphics[width=\linewidth]{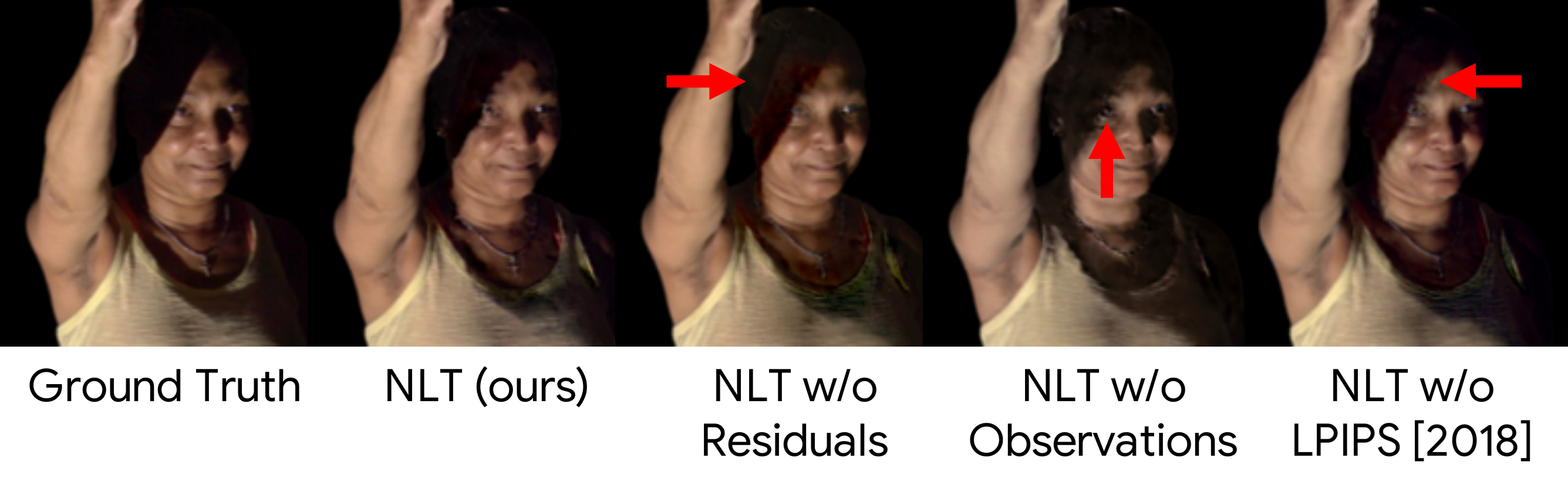}
  \vspace{-0.7cm}
  \caption{\textbf{Ablation studies} in the context of relighting.
  Removing different components of our model reduces rendering quality: no direct access to the diffuse base makes it more challenging for the network to learn hard shadows, having no observation path deprives the network of information from nearby views or lights, and removing the perceptual loss of \citet{zhang_unreasonable_2018} blurs the shadow boundary.
  }
  \label{fig:ablation}
\end{figure}

\myparagraph{No observation paths.} 
Instead of our two-path query/observation network (Section~\ref{sec:queryobs}) we can just train the query path of our network without the available observations. As shown in Figure~\ref{fig:ablation}, this ablation struggles to synthesize details for each possible view and lighting condition, and produces oversmoothed results.

\myparagraph{No residual learning.} Instead of using our residual learning approach (Section~\ref{sec:residual}) we can allow our network to directly predict the output image.
As shown in Figure~\ref{fig:ablation},
\rev{not using the diffuse base at all}
reduces the quality of the rendered image, likely because the network is then forced to waste its capacity on inferring shadows and albedo.
\rev{The middle ground between no diffuse base and our full method is using the diffuse bases only as network input, but not for the skip link. Comparing the ``Deep Shading'' rows and ``NLT w/o obs.'' rows of \tbl{tbl:relight_dir} and \tbl{tbl:viewsyn} reveals the importance of the skip connection to diffuse bases: in both relighting and view synthesis, NLT without observations (which has the skip link) outperforms Deep Shading (which uses the diffuse bases only as network input) in LPIPS.}
Our proposed residual learning scheme allows our model to focus on learning higher-order light transport effects, which results in more realistic renderings.

\myparagraph{No perceptual loss}
We find that adding a perceptual loss as proposed by~\citet{zhang_unreasonable_2018} helps the network produce higher-frequency details (such as the hard shadow boundary in~\fig{fig:ablation}). Quantitative evaluations verify this observation: full \model with the perceptual loss achieves the best perceptual scores in both tasks of relighting and view synthesis.

\section{Limitations}
\label{sec:limitations}

Similar to recent works in neural rendering \cite{neural_volumes,thies_deferred_2019,lombardi_deep_2018,sitzmann_deepvoxels:_2019,mildenhall2020nerf}, our method must be trained individually per scene, and generalizing to unseen scenes is an important future step for the field. \rev{In addition, neural rendering of dynamic scenes is desirable, especially in this case of human subjects. Using a fixed texture atlas may directly enable our method to work for dynamic performers.} 

Additionally, the fixed $1024 \times 1024$ resolution of our texture-space model limits our model's ability to synthesize higher-frequency contents, especially when the camera zooms very close to the subject\rev{, or when an image patch is allocated too few texels (see the hair artifact in~\fig{fig:viewsyn}D)}. This could be solved by training on higher-resolution images, but this would increase memory requirements and likely require significant engineering effort.

\begin{figure}[!htbp]

  \vspace{-0.1cm}
\centering
\includegraphics[width=\linewidth]{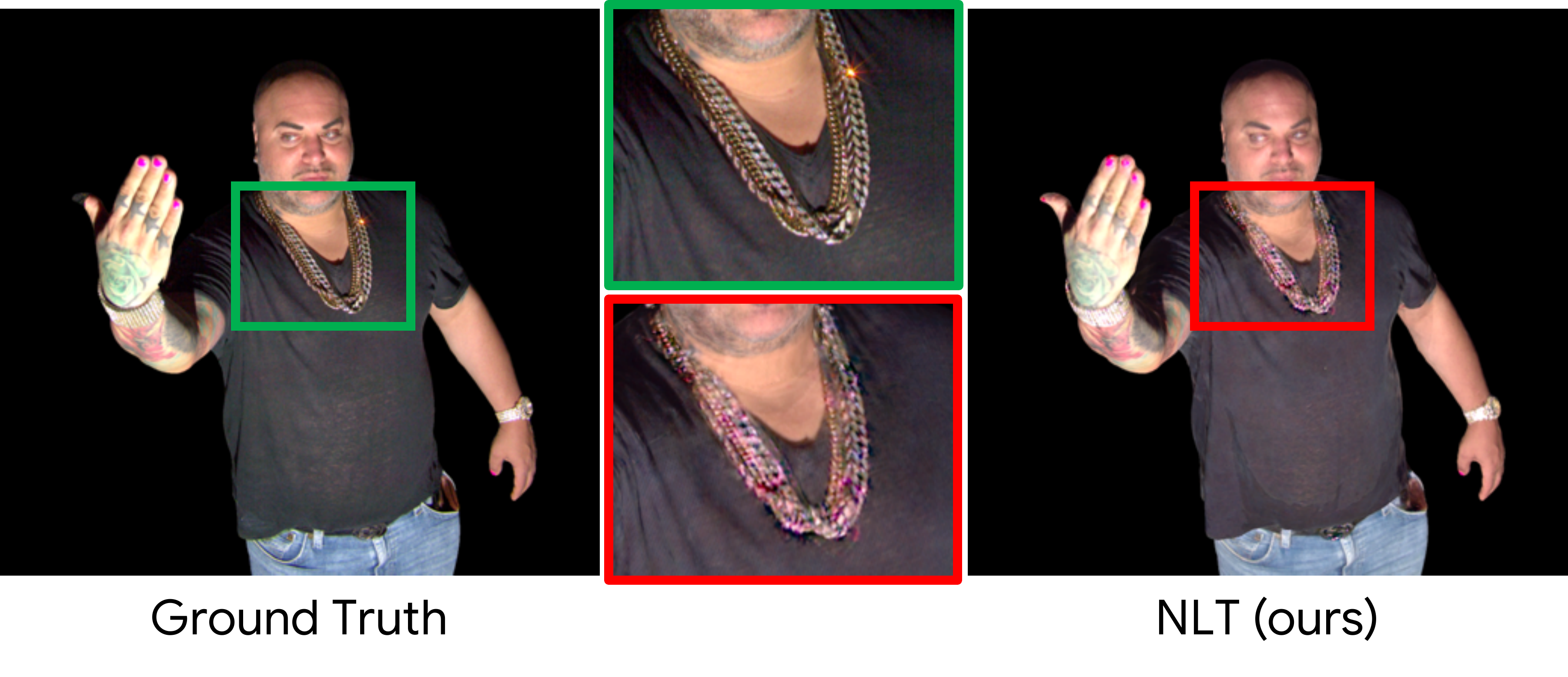}
\vspace{-0.75cm}
    \caption{\textbf{A failure case} in view synthesis. \model may fail to synthesize views of complicated light transport effects, such as those on the glittery chain.
    }
  \label{fig:failure}
  \vspace{-0.4cm}
\end{figure}

Our method has occasional failure modes as shown in Figure~\ref{fig:failure}, where complex light transport effects, such as the ones on the glittery chain, are hard to synthesize, and the final renderings lack high-frequency details.

\section{Conclusion}
\label{sec:conclu}

We have presented \nltfull, a semi-parametric deep learning framework that allows for simultaneous relighting and view synthesis of full-body scans of human subjects.
Our approach is enabled by prior work~\cite{guo_relightables:_2019} that provides a method for recovering geometric models and texture atlases, and uses as input One-Light-at-A-Time (OLAT) images captured by a light stage.
Our model works by embedding a deep neural network into the UV texture space provided by a mesh and texture atlas, and then training that model to synthesize texture-space RGB images corresponding to observed light and viewing directions.
Our model consists of a dual-path neural network architecture for aggregating information from observed images and synthesizing new images, which is further enhanced through the use of augmented texture-space inputs that leverage insights from conventional graphics techniques and a residual learning scheme that allows training to focus on higher-order light transport effects such as highlights, scattering, and global illumination. Multiple comparisons and experiments demonstrate clear improvement over previous specialized relighting or view synthesis solutions, and our approach additionally enables simultaneous relighting and view synthesis.
\begin{acks}
We thank the anonymous reviewers for their valuable feedback, Dan B.\ Goldman for pointing out an image-based relighting issue, Zhoutong Zhang, Graham Fyffe, and Xuaner (Cecilia) Zhang for the fruitful discussions, Qiurui He, Charles Herrmann, and Hayato Ikoma for the generous infrastructure support, David E.\ Jacobs and Marc Levoy for their constructive comments on an initial draft of this paper, and the actors for appearing in this paper. This work was funded in part by a Google Fellowship, ONR grant N000142012529, and the Ronald L. Graham Chair. We acknowledge support from Shell Research.
\end{acks} 
\bibliographystyle{ACM-Reference-Format}
\bibliography{main}


\begin{thebibliography}{78}


\ifx \showCODEN    \undefined \def \showCODEN     #1{\unskip}     \fi
\ifx \showDOI      \undefined \def \showDOI       #1{#1}\fi
\ifx \showISBNx    \undefined \def \showISBNx     #1{\unskip}     \fi
\ifx \showISBNxiii \undefined \def \showISBNxiii  #1{\unskip}     \fi
\ifx \showISSN     \undefined \def \showISSN      #1{\unskip}     \fi
\ifx \showLCCN     \undefined \def \showLCCN      #1{\unskip}     \fi
\ifx \shownote     \undefined \def \shownote      #1{#1}          \fi
\ifx \showarticletitle \undefined \def \showarticletitle #1{#1}   \fi
\ifx \showURL      \undefined \def \showURL       {\relax}        \fi
\providecommand\bibfield[2]{#2}
\providecommand\bibinfo[2]{#2}
\providecommand\natexlab[1]{#1}
\providecommand\showeprint[2][]{arXiv:#2}

\bibitem[\protect\citeauthoryear{Abadi, Barham, Chen, Chen, Davis, Dean, Devin,
  Ghemawat, Irving, Isard, and et~al.}{Abadi et~al\mbox{.}}{2016}]%
        {tensorflow}
\bibfield{author}{\bibinfo{person}{Mart\'{\i}n Abadi}, \bibinfo{person}{Paul
  Barham}, \bibinfo{person}{Jianmin Chen}, \bibinfo{person}{Zhifeng Chen},
  \bibinfo{person}{Andy Davis}, \bibinfo{person}{Jeffrey Dean},
  \bibinfo{person}{Matthieu Devin}, \bibinfo{person}{Sanjay Ghemawat},
  \bibinfo{person}{Geoffrey Irving}, \bibinfo{person}{Michael Isard}, {and}
  \bibinfo{person}{et al.}} \bibinfo{year}{2016}\natexlab{}.
\newblock \showarticletitle{TensorFlow: A System for Large-Scale Machine
  Learning}. In \bibinfo{booktitle}{\emph{OSDI}}.
\newblock


\bibitem[\protect\citeauthoryear{Adelson and Bergen}{Adelson and
  Bergen}{1991}]%
        {adelson1991plenoptic}
\bibfield{author}{\bibinfo{person}{Edward~H Adelson} {and}
  \bibinfo{person}{James~R Bergen}.} \bibinfo{year}{1991}\natexlab{}.
\newblock \showarticletitle{The plenoptic function and the elements of early
  vision}.
\newblock \bibinfo{journal}{\emph{Computational Models of Visual Processing}}
  (\bibinfo{year}{1991}).
\newblock


\bibitem[\protect\citeauthoryear{Alldieck, Pons-Moll, Theobalt, and
  Magnor}{Alldieck et~al\mbox{.}}{2019}]%
        {alldieck2019tex2shape}
\bibfield{author}{\bibinfo{person}{Thiemo Alldieck}, \bibinfo{person}{Gerard
  Pons-Moll}, \bibinfo{person}{Christian Theobalt}, {and}
  \bibinfo{person}{Marcus Magnor}.} \bibinfo{year}{2019}\natexlab{}.
\newblock \showarticletitle{Tex2Shape: Detailed Full Human Body Geometry from a
  Single Image}. In \bibinfo{booktitle}{\emph{ICCV}}.
\newblock


\bibitem[\protect\citeauthoryear{Barron}{Barron}{2019}]%
        {barron2019general}
\bibfield{author}{\bibinfo{person}{Jonathan~T Barron}.}
  \bibinfo{year}{2019}\natexlab{}.
\newblock \showarticletitle{A General and Adaptive Robust Loss Function}. In
  \bibinfo{booktitle}{\emph{CVPR}}.
\newblock


\bibitem[\protect\citeauthoryear{Barron and Malik}{Barron and Malik}{2015}]%
        {barron_shape_2015}
\bibfield{author}{\bibinfo{person}{Jonathan~T. Barron} {and}
  \bibinfo{person}{Jitendra Malik}.} \bibinfo{year}{2015}\natexlab{}.
\newblock \showarticletitle{Shape, {Illumination}, and {Reflectance} from
  {Shading}}.
\newblock \bibinfo{journal}{\emph{TPAMI}} (\bibinfo{year}{2015}).
\newblock


\bibitem[\protect\citeauthoryear{Barrow and Tenenbaum}{Barrow and
  Tenenbaum}{1978}]%
        {Barrow1978}
\bibfield{author}{\bibinfo{person}{H.~G. Barrow} {and} \bibinfo{person}{J.~M.
  Tenenbaum}.} \bibinfo{year}{1978}\natexlab{}.
\newblock \showarticletitle{Recovering intrinsic scene characteristics from
  images}.
\newblock \bibinfo{journal}{\emph{Computer Vision Systems}}
  (\bibinfo{year}{1978}).
\newblock


\bibitem[\protect\citeauthoryear{Basri, Jacobs, and Kemelmacher}{Basri
  et~al\mbox{.}}{2007}]%
        {basri2007photometric}
\bibfield{author}{\bibinfo{person}{Ronen Basri}, \bibinfo{person}{David
  Jacobs}, {and} \bibinfo{person}{Ira Kemelmacher}.}
  \bibinfo{year}{2007}\natexlab{}.
\newblock \showarticletitle{Photometric stereo with general, unknown lighting}.
\newblock \bibinfo{journal}{\emph{IJCV}} (\bibinfo{year}{2007}).
\newblock


\bibitem[\protect\citeauthoryear{Buehler, Bosse, McMillan, Gortler, and
  Cohen}{Buehler et~al\mbox{.}}{2001}]%
        {buehler2001unstructured}
\bibfield{author}{\bibinfo{person}{Chris Buehler}, \bibinfo{person}{Michael
  Bosse}, \bibinfo{person}{Leonard McMillan}, \bibinfo{person}{Steven Gortler},
  {and} \bibinfo{person}{Michael Cohen}.} \bibinfo{year}{2001}\natexlab{}.
\newblock \showarticletitle{Unstructured lumigraph rendering}. In
  \bibinfo{booktitle}{\emph{Proceedings of the 28th annual conference on
  Computer graphics and interactive techniques}}. \bibinfo{pages}{425--432}.
\newblock


\bibitem[\protect\citeauthoryear{Carranza, Theobalt, Magnor, and
  Seidel}{Carranza et~al\mbox{.}}{2003}]%
        {carranza2003free}
\bibfield{author}{\bibinfo{person}{Joel Carranza}, \bibinfo{person}{Christian
  Theobalt}, \bibinfo{person}{Marcus~A Magnor}, {and}
  \bibinfo{person}{Hans-Peter Seidel}.} \bibinfo{year}{2003}\natexlab{}.
\newblock \showarticletitle{Free-viewpoint video of human actors}. In
  \bibinfo{booktitle}{\emph{ACM SIGGRAPH}}, Vol.~\bibinfo{volume}{22}.
  \bibinfo{pages}{569--577}.
\newblock


\bibitem[\protect\citeauthoryear{Chen, Chen, Zhang, Wang, Ji, Kutulakos, and
  Yu}{Chen et~al\mbox{.}}{2020}]%
        {chen2020neural}
\bibfield{author}{\bibinfo{person}{Zhang Chen}, \bibinfo{person}{Anpei Chen},
  \bibinfo{person}{Guli Zhang}, \bibinfo{person}{Chengyuan Wang},
  \bibinfo{person}{Yu Ji}, \bibinfo{person}{Kiriakos~N Kutulakos}, {and}
  \bibinfo{person}{Jingyi Yu}.} \bibinfo{year}{2020}\natexlab{}.
\newblock \showarticletitle{A Neural Rendering Framework for Free-Viewpoint
  Relighting}. In \bibinfo{booktitle}{\emph{Proceedings of the IEEE/CVF
  Conference on Computer Vision and Pattern Recognition}}.
  \bibinfo{pages}{5599--5610}.
\newblock


\bibitem[\protect\citeauthoryear{Cohen, Daubechies, and Feauveau}{Cohen
  et~al\mbox{.}}{1992}]%
        {wavelet}
\bibfield{author}{\bibinfo{person}{A. Cohen}, \bibinfo{person}{Ingrid
  Daubechies}, {and} \bibinfo{person}{J.-C. Feauveau}.}
  \bibinfo{year}{1992}\natexlab{}.
\newblock \showarticletitle{Biorthogonal Bases of Compactly Supported
  Wavelets}.
\newblock \bibinfo{journal}{\emph{Communications on Pure and Applied
  Mathematics}} (\bibinfo{year}{1992}).
\newblock


\bibitem[\protect\citeauthoryear{Collet, Chuang, Sweeney, Gillett, Evseev,
  Calabrese, Hoppe, Kirk, and Sullivan}{Collet et~al\mbox{.}}{2015}]%
        {fvv}
\bibfield{author}{\bibinfo{person}{Alvaro Collet}, \bibinfo{person}{Ming
  Chuang}, \bibinfo{person}{Pat Sweeney}, \bibinfo{person}{Don Gillett},
  \bibinfo{person}{Dennis Evseev}, \bibinfo{person}{David Calabrese},
  \bibinfo{person}{Hugues Hoppe}, \bibinfo{person}{Adam Kirk}, {and}
  \bibinfo{person}{Steve Sullivan}.} \bibinfo{year}{2015}\natexlab{}.
\newblock \showarticletitle{High-quality Streamable Free-viewpoint Video}.
\newblock \bibinfo{journal}{\emph{ACM TOG}} (\bibinfo{year}{2015}).
\newblock


\bibitem[\protect\citeauthoryear{Debevec}{Debevec}{2012}]%
        {debevec2012light}
\bibfield{author}{\bibinfo{person}{Paul Debevec}.}
  \bibinfo{year}{2012}\natexlab{}.
\newblock \showarticletitle{The Light Stages and Their Applications to
  Photoreal Digital Actors}.
\newblock  (\bibinfo{year}{2012}).
\newblock


\bibitem[\protect\citeauthoryear{Debevec, Hawkins, Tchou, Duiker, Sarokin, and
  Sagar}{Debevec et~al\mbox{.}}{2000}]%
        {debevec_acquiring_2000}
\bibfield{author}{\bibinfo{person}{Paul Debevec}, \bibinfo{person}{Tim
  Hawkins}, \bibinfo{person}{Chris Tchou}, \bibinfo{person}{Haarm-Pieter
  Duiker}, \bibinfo{person}{Westley Sarokin}, {and} \bibinfo{person}{Mark
  Sagar}.} \bibinfo{year}{2000}\natexlab{}.
\newblock \showarticletitle{Acquiring the reflectance field of a human face}.
  In \bibinfo{booktitle}{\emph{SIGGRAPH}}.
\newblock


\bibitem[\protect\citeauthoryear{Deering, Winner, Schediwy, Duffy, and
  Hunt}{Deering et~al\mbox{.}}{1988}]%
        {gbuffer}
\bibfield{author}{\bibinfo{person}{Michael Deering}, \bibinfo{person}{Stephanie
  Winner}, \bibinfo{person}{Bic Schediwy}, \bibinfo{person}{Chris Duffy}, {and}
  \bibinfo{person}{Neil Hunt}.} \bibinfo{year}{1988}\natexlab{}.
\newblock \showarticletitle{The Triangle Processor and Normal Vector Shader: A
  VLSI System for High Performance Graphics}. In
  \bibinfo{booktitle}{\emph{SIGGRAPH}}.
\newblock


\bibitem[\protect\citeauthoryear{Deng, Dong, Socher, Li, Li, and Fei-Fei}{Deng
  et~al\mbox{.}}{2009}]%
        {imagenet_cvpr09}
\bibfield{author}{\bibinfo{person}{J. Deng}, \bibinfo{person}{W. Dong},
  \bibinfo{person}{R. Socher}, \bibinfo{person}{L.-J. Li}, \bibinfo{person}{K.
  Li}, {and} \bibinfo{person}{L. Fei-Fei}.} \bibinfo{year}{2009}\natexlab{}.
\newblock \showarticletitle{{ImageNet: A Large-Scale Hierarchical Image
  Database}}. In \bibinfo{booktitle}{\emph{CVPR09}}.
\newblock


\bibitem[\protect\citeauthoryear{Eigen, Puhrsch, and Fergus}{Eigen
  et~al\mbox{.}}{2014}]%
        {eigen2014depth}
\bibfield{author}{\bibinfo{person}{David Eigen}, \bibinfo{person}{Christian
  Puhrsch}, {and} \bibinfo{person}{Rob Fergus}.}
  \bibinfo{year}{2014}\natexlab{}.
\newblock \showarticletitle{Depth Map Prediction from a Single Image Using a
  Multi-Scale Deep Network}. In \bibinfo{booktitle}{\emph{NIPS}}.
\newblock


\bibitem[\protect\citeauthoryear{Eisemann, De~Decker, Magnor, Bekaert,
  De~Aguiar, Ahmed, Theobalt, and Sellent}{Eisemann et~al\mbox{.}}{2008}]%
        {eisemann2008floating}
\bibfield{author}{\bibinfo{person}{Martin Eisemann}, \bibinfo{person}{Bert
  De~Decker}, \bibinfo{person}{Marcus Magnor}, \bibinfo{person}{Philippe
  Bekaert}, \bibinfo{person}{Edilson De~Aguiar}, \bibinfo{person}{Naveed
  Ahmed}, \bibinfo{person}{Christian Theobalt}, {and} \bibinfo{person}{Anita
  Sellent}.} \bibinfo{year}{2008}\natexlab{}.
\newblock \showarticletitle{Floating textures}. In
  \bibinfo{booktitle}{\emph{Computer graphics forum}}.
\newblock


\bibitem[\protect\citeauthoryear{Flynn, Broxton, Debevec, DuVall, Fyffe,
  Overbeck, Snavely, and Tucker}{Flynn et~al\mbox{.}}{2019}]%
        {Flynn_2019_CVPR}
\bibfield{author}{\bibinfo{person}{John Flynn}, \bibinfo{person}{Michael
  Broxton}, \bibinfo{person}{Paul Debevec}, \bibinfo{person}{Matthew DuVall},
  \bibinfo{person}{Graham Fyffe}, \bibinfo{person}{Ryan Overbeck},
  \bibinfo{person}{Noah Snavely}, {and} \bibinfo{person}{Richard Tucker}.}
  \bibinfo{year}{2019}\natexlab{}.
\newblock \showarticletitle{DeepView: View Synthesis With Learned Gradient
  Descent}. In \bibinfo{booktitle}{\emph{CVPR}}.
\newblock


\bibitem[\protect\citeauthoryear{Fyffe, Wilson, and Debevec}{Fyffe
  et~al\mbox{.}}{2009}]%
        {Fyffe:2009}
\bibfield{author}{\bibinfo{person}{Graham Fyffe}, \bibinfo{person}{Cyrus~A.
  Wilson}, {and} \bibinfo{person}{Paul Debevec}.}
  \bibinfo{year}{2009}\natexlab{}.
\newblock \showarticletitle{Cosine Lobe Based Relighting from Gradient
  Illumination Photographs}. In \bibinfo{booktitle}{\emph{SIGGRAPH Poster}}.
\newblock


\bibitem[\protect\citeauthoryear{Gardner, Hold-Geoffroy, Sunkavalli, Gagne, and
  Lalonde}{Gardner et~al\mbox{.}}{2019}]%
        {gardner_deep_nodate}
\bibfield{author}{\bibinfo{person}{Marc-Andre Gardner},
  \bibinfo{person}{Yannick Hold-Geoffroy}, \bibinfo{person}{Kalyan Sunkavalli},
  \bibinfo{person}{Christian Gagne}, {and} \bibinfo{person}{Jean-Francois
  Lalonde}.} \bibinfo{year}{2019}\natexlab{}.
\newblock \showarticletitle{Deep {Parametric} {Indoor} {Lighting}
  {Estimation}}.
\newblock  (\bibinfo{year}{2019}).
\newblock


\bibitem[\protect\citeauthoryear{Garg, Talvala, Levoy, and Lensch}{Garg
  et~al\mbox{.}}{2006}]%
        {garg2006symmetric}
\bibfield{author}{\bibinfo{person}{Gaurav Garg}, \bibinfo{person}{Eino-Ville
  Talvala}, \bibinfo{person}{Marc Levoy}, {and} \bibinfo{person}{Hendrik~PA
  Lensch}.} \bibinfo{year}{2006}\natexlab{}.
\newblock \showarticletitle{Symmetric Photography: Exploiting Data-sparseness
  in Reflectance Fields}. In \bibinfo{booktitle}{\emph{Rendering Techniques}}.
  \bibinfo{pages}{251--262}.
\newblock


\bibitem[\protect\citeauthoryear{Gortler, Grzeszczuk, Szeliski, and
  Cohen}{Gortler et~al\mbox{.}}{1996}]%
        {Gortler}
\bibfield{author}{\bibinfo{person}{S Gortler}, \bibinfo{person}{R Grzeszczuk},
  \bibinfo{person}{R Szeliski}, {and} \bibinfo{person}{M Cohen}.}
  \bibinfo{year}{1996}\natexlab{}.
\newblock \showarticletitle{The Lumigraph}. In
  \bibinfo{booktitle}{\emph{SIGGRAPH}}.
\newblock


\bibitem[\protect\citeauthoryear{Guo, Dourgarian, Tang, tkach, Kowdle, Cooper,
  Dou, Fanello, Fyffe, Rhemann, Taylor, Lincoln, Debevec, Izad, Davidson,
  Busch, Yu, Whalen, Harvey, Orts-Escolano, and Pandey}{Guo
  et~al\mbox{.}}{2019}]%
        {guo_relightables:_2019}
\bibfield{author}{\bibinfo{person}{Kaiwen Guo}, \bibinfo{person}{Jason
  Dourgarian}, \bibinfo{person}{Danhang Tang}, \bibinfo{person}{Anastasia
  tkach}, \bibinfo{person}{Adarsh Kowdle}, \bibinfo{person}{Emily Cooper},
  \bibinfo{person}{Mingsong Dou}, \bibinfo{person}{Sean Fanello},
  \bibinfo{person}{Graham Fyffe}, \bibinfo{person}{Christoph Rhemann},
  \bibinfo{person}{Jonathan Taylor}, \bibinfo{person}{Peter Lincoln},
  \bibinfo{person}{Paul Debevec}, \bibinfo{person}{Shahram Izad},
  \bibinfo{person}{Philip Davidson}, \bibinfo{person}{Jay Busch},
  \bibinfo{person}{Xueming Yu}, \bibinfo{person}{Matt Whalen},
  \bibinfo{person}{Geoff Harvey}, \bibinfo{person}{Sergio Orts-Escolano}, {and}
  \bibinfo{person}{Rohit Pandey}.} \bibinfo{year}{2019}\natexlab{}.
\newblock \showarticletitle{The {R}elightables: {V}olumetric {P}erformance
  {C}apture of {H}umans with {R}ealistic {R}elighting}. In
  \bibinfo{booktitle}{\emph{SIGGRAPH Asia}}.
\newblock


\bibitem[\protect\citeauthoryear{Hartley and Zisserman}{Hartley and
  Zisserman}{2003}]%
        {Hartley2003}
\bibfield{author}{\bibinfo{person}{Richard Hartley} {and}
  \bibinfo{person}{Andrew Zisserman}.} \bibinfo{year}{2003}\natexlab{}.
\newblock \bibinfo{booktitle}{\emph{Multiple View Geometry in Computer Vision}
  (\bibinfo{edition}{2} ed.)}.
\newblock \bibinfo{publisher}{Cambridge University Press},
  \bibinfo{address}{New York, NY, USA}.
\newblock
\showISBNx{0521540518}


\bibitem[\protect\citeauthoryear{He, Zhang, Ren, and Sun}{He
  et~al\mbox{.}}{2016}]%
        {he2016deep}
\bibfield{author}{\bibinfo{person}{Kaiming He}, \bibinfo{person}{Xiangyu
  Zhang}, \bibinfo{person}{Shaoqing Ren}, {and} \bibinfo{person}{Jian Sun}.}
  \bibinfo{year}{2016}\natexlab{}.
\newblock \showarticletitle{Deep residual learning for image recognition}.
\newblock \bibinfo{journal}{\emph{CVPR}} (\bibinfo{year}{2016}).
\newblock


\bibitem[\protect\citeauthoryear{Hornik}{Hornik}{1991}]%
        {hornik1991approximation}
\bibfield{author}{\bibinfo{person}{Kurt Hornik}.}
  \bibinfo{year}{1991}\natexlab{}.
\newblock \showarticletitle{Approximation Capabilities of Multilayer
  Feedforward Networks}.
\newblock \bibinfo{journal}{\emph{Neural Networks}} (\bibinfo{year}{1991}).
\newblock


\bibitem[\protect\citeauthoryear{Kajiya}{Kajiya}{1986}]%
        {Kajiya_rendeq_1986}
\bibfield{author}{\bibinfo{person}{James~T. Kajiya}.}
  \bibinfo{year}{1986}\natexlab{}.
\newblock \showarticletitle{The Rendering Equation}. In
  \bibinfo{booktitle}{\emph{SIGGRAPH}}.
\newblock


\bibitem[\protect\citeauthoryear{Kanamori and Endo}{Kanamori and Endo}{2018}]%
        {relightinghumans}
\bibfield{author}{\bibinfo{person}{Yoshihiro Kanamori} {and}
  \bibinfo{person}{Yuki Endo}.} \bibinfo{year}{2018}\natexlab{}.
\newblock \showarticletitle{Relighting Humans: Occlusion-Aware Inverse
  Rendering for Full-Body Human Images}.
\newblock \bibinfo{journal}{\emph{TOG}} (\bibinfo{year}{2018}).
\newblock


\bibitem[\protect\citeauthoryear{Kelly, Cordingley, Nolan, Rhemann, Fanello,
  Tang, Osborn, Busch, Davidson, Debevec, Denny, Fyffe, Guo, Harvey, Izadi,
  Lincoln, Ma, Taylor, Yu, Whalen, Dourgarian, Blanchett, French, Sillitoe,
  Uglow, Spiteri, Pearson, Kernot, and Richards}{Kelly et~al\mbox{.}}{2019}]%
        {aria}
\bibfield{author}{\bibinfo{person}{Sean Kelly}, \bibinfo{person}{Samantha
  Cordingley}, \bibinfo{person}{Patrick Nolan}, \bibinfo{person}{Christoph
  Rhemann}, \bibinfo{person}{Sean Fanello}, \bibinfo{person}{Danhang Tang},
  \bibinfo{person}{Jude Osborn}, \bibinfo{person}{Jay Busch},
  \bibinfo{person}{Philip Davidson}, \bibinfo{person}{Paul Debevec},
  \bibinfo{person}{Peter Denny}, \bibinfo{person}{Graham Fyffe},
  \bibinfo{person}{Kaiwen Guo}, \bibinfo{person}{Geoff Harvey},
  \bibinfo{person}{Shahram Izadi}, \bibinfo{person}{Peter Lincoln},
  \bibinfo{person}{Wan-Chun~Alex Ma}, \bibinfo{person}{Jonathan Taylor},
  \bibinfo{person}{Xueming Yu}, \bibinfo{person}{Matt Whalen},
  \bibinfo{person}{Jason Dourgarian}, \bibinfo{person}{Genevieve Blanchett},
  \bibinfo{person}{Narelle French}, \bibinfo{person}{Kirstin Sillitoe},
  \bibinfo{person}{Tea Uglow}, \bibinfo{person}{Brenton Spiteri},
  \bibinfo{person}{Emma Pearson}, \bibinfo{person}{Wade Kernot}, {and}
  \bibinfo{person}{Jonathan Richards}.} \bibinfo{year}{2019}\natexlab{}.
\newblock \showarticletitle{AR-ia: Volumetric Opera for Mobile Augmented
  Reality}. In \bibinfo{booktitle}{\emph{SIGGRAPH Asia 2019 XR}}.
\newblock


\bibitem[\protect\citeauthoryear{Kim, Garrido, Tewari, Xu, Thies, Nie{\ss}ner,
  P{\'e}rez, Richardt, Zoll{\"o}fer, and Theobalt}{Kim et~al\mbox{.}}{2018}]%
        {kim2018deep}
\bibfield{author}{\bibinfo{person}{Hyeongwoo Kim}, \bibinfo{person}{Pablo
  Garrido}, \bibinfo{person}{Ayush Tewari}, \bibinfo{person}{Weipeng Xu},
  \bibinfo{person}{Justus Thies}, \bibinfo{person}{Matthias Nie{\ss}ner},
  \bibinfo{person}{Patrick P{\'e}rez}, \bibinfo{person}{Christian Richardt},
  \bibinfo{person}{Michael Zoll{\"o}fer}, {and} \bibinfo{person}{Christian
  Theobalt}.} \bibinfo{year}{2018}\natexlab{}.
\newblock \showarticletitle{Deep Video Portraits}.
\newblock \bibinfo{journal}{\emph{TOG}} (\bibinfo{year}{2018}).
\newblock


\bibitem[\protect\citeauthoryear{Kingma and Ba}{Kingma and Ba}{2015}]%
        {KingmaB15}
\bibfield{author}{\bibinfo{person}{Diederik~P. Kingma} {and}
  \bibinfo{person}{Jimmy Ba}.} \bibinfo{year}{2015}\natexlab{}.
\newblock \showarticletitle{Adam: A Method for Stochastic Optimization}. In
  \bibinfo{booktitle}{\emph{ICLR}}.
\newblock


\bibitem[\protect\citeauthoryear{LeGendre, Ma, Fyffe, Flynn, Charbonnel, Busch,
  and Debevec}{LeGendre et~al\mbox{.}}{2019}]%
        {legendre2019deeplight}
\bibfield{author}{\bibinfo{person}{Chloe LeGendre}, \bibinfo{person}{Wan-Chun
  Ma}, \bibinfo{person}{Graham Fyffe}, \bibinfo{person}{John Flynn},
  \bibinfo{person}{Laurent Charbonnel}, \bibinfo{person}{Jay Busch}, {and}
  \bibinfo{person}{Paul Debevec}.} \bibinfo{year}{2019}\natexlab{}.
\newblock \showarticletitle{DeepLight: Learning Illumination for Unconstrained
  Mobile Mixed Reality}. In \bibinfo{booktitle}{\emph{CVPR}}.
\newblock


\bibitem[\protect\citeauthoryear{Levoy and Hanrahan}{Levoy and
  Hanrahan}{1996}]%
        {lightfield}
\bibfield{author}{\bibinfo{person}{Marc Levoy} {and} \bibinfo{person}{Pat
  Hanrahan}.} \bibinfo{year}{1996}\natexlab{}.
\newblock \showarticletitle{Light Field Rendering}. In
  \bibinfo{booktitle}{\emph{SIGGRAPH}}.
\newblock


\bibitem[\protect\citeauthoryear{Li, Mao, Zhang, Freeman, Tenenbaum, Snavely,
  and Wu}{Li et~al\mbox{.}}{2020}]%
        {li2020multi}
\bibfield{author}{\bibinfo{person}{Yikai Li}, \bibinfo{person}{Jiayuan Mao},
  \bibinfo{person}{Xiuming Zhang}, \bibinfo{person}{Bill Freeman},
  \bibinfo{person}{Josh Tenenbaum}, \bibinfo{person}{Noah Snavely}, {and}
  \bibinfo{person}{Jiajun Wu}.} \bibinfo{year}{2020}\natexlab{}.
\newblock \showarticletitle{Multi-Plane Program Induction with 3D Box Priors}.
\newblock \bibinfo{journal}{\emph{Advances in Neural Information Processing
  Systems}}  \bibinfo{volume}{33} (\bibinfo{year}{2020}).
\newblock


\bibitem[\protect\citeauthoryear{Li, Wiedemann, and Mitchell}{Li
  et~al\mbox{.}}{2019}]%
        {li19}
\bibfield{author}{\bibinfo{person}{Yue Li}, \bibinfo{person}{Pablo Wiedemann},
  {and} \bibinfo{person}{Kenny Mitchell}.} \bibinfo{year}{2019}\natexlab{}.
\newblock \showarticletitle{Deep Precomputed Radiance Transfer for Deformable
  Objects}.
\newblock \bibinfo{journal}{\emph{ACM CGIT}} (\bibinfo{year}{2019}).
\newblock


\bibitem[\protect\citeauthoryear{Li, Xu, Ramamoorthi, Sunkavalli, and
  Chandraker}{Li et~al\mbox{.}}{2018}]%
        {Zhengqin2}
\bibfield{author}{\bibinfo{person}{Z. Li}, \bibinfo{person}{Z. Xu},
  \bibinfo{person}{R. Ramamoorthi}, \bibinfo{person}{K. Sunkavalli}, {and}
  \bibinfo{person}{M. Chandraker}.} \bibinfo{year}{2018}\natexlab{}.
\newblock \showarticletitle{Learning to Reconstruct Shape and Spatially-Varying
  Reflectance from a Single Image}. In \bibinfo{booktitle}{\emph{SIGGRAPH
  Asia}}.
\newblock


\bibitem[\protect\citeauthoryear{Lombardi, Saragih, Simon, and Sheikh}{Lombardi
  et~al\mbox{.}}{2018}]%
        {lombardi_deep_2018}
\bibfield{author}{\bibinfo{person}{Stephen Lombardi}, \bibinfo{person}{Jason
  Saragih}, \bibinfo{person}{Tomas Simon}, {and} \bibinfo{person}{Yaser
  Sheikh}.} \bibinfo{year}{2018}\natexlab{}.
\newblock \showarticletitle{Deep {Appearance} {Models} for {Face} {Rendering}}.
\newblock \bibinfo{journal}{\emph{ACM TOG}} (\bibinfo{year}{2018}).
\newblock


\bibitem[\protect\citeauthoryear{Lombardi, Simon, Saragih, Schwartz, Lehrmann,
  and Sheikh}{Lombardi et~al\mbox{.}}{2019}]%
        {neural_volumes}
\bibfield{author}{\bibinfo{person}{Stephen Lombardi}, \bibinfo{person}{Tomas
  Simon}, \bibinfo{person}{Jason Saragih}, \bibinfo{person}{Gabriel Schwartz},
  \bibinfo{person}{Andreas Lehrmann}, {and} \bibinfo{person}{Yaser Sheikh}.}
  \bibinfo{year}{2019}\natexlab{}.
\newblock \showarticletitle{Neural Volumes: Learning Dynamic Renderable Volumes
  from Images}.
\newblock \bibinfo{journal}{\emph{SIGGRAPH}} (\bibinfo{year}{2019}).
\newblock


\bibitem[\protect\citeauthoryear{Ma, Hawkins, Peers, Chabert, Weiss, and
  Debevec}{Ma et~al\mbox{.}}{2007}]%
        {Ma:2007}
\bibfield{author}{\bibinfo{person}{Wan-Chun Ma}, \bibinfo{person}{Tim Hawkins},
  \bibinfo{person}{Pieter Peers}, \bibinfo{person}{Charles-Felix Chabert},
  \bibinfo{person}{Malte Weiss}, {and} \bibinfo{person}{Paul Debevec}.}
  \bibinfo{year}{2007}\natexlab{}.
\newblock \showarticletitle{Rapid Acquisition of Specular and Diffuse Normal
  Maps from Polarized Spherical Gradient Illumination}. In
  \bibinfo{booktitle}{\emph{EGSR}}.
\newblock


\bibitem[\protect\citeauthoryear{Maas, Hannun, and Ng}{Maas
  et~al\mbox{.}}{2013}]%
        {maas2013rectifier}
\bibfield{author}{\bibinfo{person}{Andrew~L Maas}, \bibinfo{person}{Awni~Y
  Hannun}, {and} \bibinfo{person}{Andrew~Y Ng}.}
  \bibinfo{year}{2013}\natexlab{}.
\newblock \showarticletitle{Rectifier Nonlinearities Improve Neural Network
  Acoustic Models}. In \bibinfo{booktitle}{\emph{ICML}}.
\newblock


\bibitem[\protect\citeauthoryear{Martin-Brualla, Pandey, Yang, Pidlypenskyi,
  Taylor, Valentin, Khamis, Davidson, Tkach, Lincoln, and
  et~al.}{Martin-Brualla et~al\mbox{.}}{2018}]%
        {lookinggood}
\bibfield{author}{\bibinfo{person}{Ricardo Martin-Brualla},
  \bibinfo{person}{Rohit Pandey}, \bibinfo{person}{Shuoran Yang},
  \bibinfo{person}{Pavel Pidlypenskyi}, \bibinfo{person}{Jonathan Taylor},
  \bibinfo{person}{Julien Valentin}, \bibinfo{person}{Sameh Khamis},
  \bibinfo{person}{Philip Davidson}, \bibinfo{person}{Anastasia Tkach},
  \bibinfo{person}{Peter Lincoln}, {and} \bibinfo{person}{et al.}}
  \bibinfo{year}{2018}\natexlab{}.
\newblock \showarticletitle{LookinGood: Enhancing Performance Capture with
  Real-Time Neural Re-Rendering}.
\newblock \bibinfo{journal}{\emph{SIGGRAPH Asia}}.
\newblock


\bibitem[\protect\citeauthoryear{Meka, Dourgarian, Denny, Bouaziz, Lincoln,
  Whalen, Harvey, Taylor, Izadi, Tagliasacchi, Debevec, Häne, Theobalt,
  Valentin, Rhemann, Pandey, Zollhöfer, Fanello, Fyffe, Kowdle, Yu, and
  Busch}{Meka et~al\mbox{.}}{2019}]%
        {meka_deep_2019}
\bibfield{author}{\bibinfo{person}{Abhimitra Meka}, \bibinfo{person}{Jason
  Dourgarian}, \bibinfo{person}{Peter Denny}, \bibinfo{person}{Sofien Bouaziz},
  \bibinfo{person}{Peter Lincoln}, \bibinfo{person}{Matt Whalen},
  \bibinfo{person}{Geoff Harvey}, \bibinfo{person}{Jonathan Taylor},
  \bibinfo{person}{Shahram Izadi}, \bibinfo{person}{Andrea Tagliasacchi},
  \bibinfo{person}{Paul Debevec}, \bibinfo{person}{Christian Häne},
  \bibinfo{person}{Christian Theobalt}, \bibinfo{person}{Julien Valentin},
  \bibinfo{person}{Christoph Rhemann}, \bibinfo{person}{Rohit Pandey},
  \bibinfo{person}{Michael Zollhöfer}, \bibinfo{person}{Sean Fanello},
  \bibinfo{person}{Graham Fyffe}, \bibinfo{person}{Adarsh Kowdle},
  \bibinfo{person}{Xueming Yu}, {and} \bibinfo{person}{Jay Busch}.}
  \bibinfo{year}{2019}\natexlab{}.
\newblock \showarticletitle{Deep reflectance fields: high-quality facial
  reflectance field inference from color gradient illumination}. In
  \bibinfo{booktitle}{\emph{SIGGRAPH}}.
\newblock


\bibitem[\protect\citeauthoryear{Mildenhall, Srinivasan, Ortiz-Cayon,
  Kalantari, Ramamoorthi, Ng, and Kar}{Mildenhall et~al\mbox{.}}{2019}]%
        {mildenhall2019llff}
\bibfield{author}{\bibinfo{person}{Ben Mildenhall}, \bibinfo{person}{Pratul~P.
  Srinivasan}, \bibinfo{person}{Rodrigo Ortiz-Cayon},
  \bibinfo{person}{Nima~Khademi Kalantari}, \bibinfo{person}{Ravi Ramamoorthi},
  \bibinfo{person}{Ren Ng}, {and} \bibinfo{person}{Abhishek Kar}.}
  \bibinfo{year}{2019}\natexlab{}.
\newblock \showarticletitle{Local Light Field Fusion: Practical View Synthesis
  with Prescriptive Sampling Guidelines}. In
  \bibinfo{booktitle}{\emph{SIGGRAPH}}.
\newblock


\bibitem[\protect\citeauthoryear{Mildenhall, Srinivasan, Tancik, Barron,
  Ramamoorthi, and Ng}{Mildenhall et~al\mbox{.}}{2020}]%
        {mildenhall2020nerf}
\bibfield{author}{\bibinfo{person}{Ben Mildenhall}, \bibinfo{person}{Pratul~P
  Srinivasan}, \bibinfo{person}{Matthew Tancik}, \bibinfo{person}{Jonathan~T
  Barron}, \bibinfo{person}{Ravi Ramamoorthi}, {and} \bibinfo{person}{Ren Ng}.}
  \bibinfo{year}{2020}\natexlab{}.
\newblock \showarticletitle{NeRF: Representing Scenes as Neural Radiance Fields
  for View Synthesis}.
\newblock \bibinfo{journal}{\emph{arXiv preprint arXiv:2003.08934}}
  (\bibinfo{year}{2020}).
\newblock


\bibitem[\protect\citeauthoryear{Murray-Coleman and Smith}{Murray-Coleman and
  Smith}{1990}]%
        {MurrayColeman_Smith}
\bibfield{author}{\bibinfo{person}{J.F. Murray-Coleman} {and}
  \bibinfo{person}{A.M. Smith}.} \bibinfo{year}{1990}\natexlab{}.
\newblock \showarticletitle{The Automated Measurement of BRDFs and their
  Application to Luminaire Modeling}.
\newblock \bibinfo{journal}{\emph{Journal of the Illuminating Engineering
  Society}} (\bibinfo{year}{1990}).
\newblock


\bibitem[\protect\citeauthoryear{Nalbach, Arabadzhiyska, Mehta, Seidel, and
  Ritschel}{Nalbach et~al\mbox{.}}{2017}]%
        {Nalbach2017b}
\bibfield{author}{\bibinfo{person}{Oliver Nalbach}, \bibinfo{person}{Elena
  Arabadzhiyska}, \bibinfo{person}{Dushyant Mehta}, \bibinfo{person}{Hans-Peter
  Seidel}, {and} \bibinfo{person}{Tobias Ritschel}.}
  \bibinfo{year}{2017}\natexlab{}.
\newblock \showarticletitle{Deep Shading: Convolutional Neural Networks for
  Screen-Space Shading}.
\newblock  (\bibinfo{year}{2017}).
\newblock


\bibitem[\protect\citeauthoryear{Nestmeyer, Lalonde, Matthews, Games, Lehrmann,
  and Borealis}{Nestmeyer et~al\mbox{.}}{2020}]%
        {nestmeyerlearning}
\bibfield{author}{\bibinfo{person}{Thomas Nestmeyer},
  \bibinfo{person}{Jean-Fran{\c{c}}ois Lalonde}, \bibinfo{person}{Iain
  Matthews}, \bibinfo{person}{Epic Games}, \bibinfo{person}{Andreas Lehrmann},
  {and} \bibinfo{person}{AI Borealis}.} \bibinfo{year}{2020}\natexlab{}.
\newblock \showarticletitle{Learning Physics-guided Face Relighting under
  Directional Light}. In \bibinfo{booktitle}{\emph{CVPR}}.
\newblock


\bibitem[\protect\citeauthoryear{Orts-Escolano, Rhemann, Fanello, Chang,
  Kowdle, Degtyarev, Kim, Davidson, Khamis, Dou, Tankovich, Loop, Cai, Chou,
  Mennicken, Valentin, Pradeep, Wang, Kang, Kohli, Lutchyn, Keskin, and
  Izadi}{Orts-Escolano et~al\mbox{.}}{2016}]%
        {holoportation}
\bibfield{author}{\bibinfo{person}{Sergio Orts-Escolano},
  \bibinfo{person}{Christoph Rhemann}, \bibinfo{person}{Sean Fanello},
  \bibinfo{person}{Wayne Chang}, \bibinfo{person}{Adarsh Kowdle},
  \bibinfo{person}{Yury Degtyarev}, \bibinfo{person}{David Kim},
  \bibinfo{person}{Philip~L. Davidson}, \bibinfo{person}{Sameh Khamis},
  \bibinfo{person}{Mingsong Dou}, \bibinfo{person}{Vladimir Tankovich},
  \bibinfo{person}{Charles Loop}, \bibinfo{person}{Qin Cai},
  \bibinfo{person}{Philip~A. Chou}, \bibinfo{person}{Sarah Mennicken},
  \bibinfo{person}{Julien Valentin}, \bibinfo{person}{Vivek Pradeep},
  \bibinfo{person}{Shenlong Wang}, \bibinfo{person}{Sing~Bing Kang},
  \bibinfo{person}{Pushmeet Kohli}, \bibinfo{person}{Yuliya Lutchyn},
  \bibinfo{person}{Cem Keskin}, {and} \bibinfo{person}{Shahram Izadi}.}
  \bibinfo{year}{2016}\natexlab{}.
\newblock \showarticletitle{Holoportation: Virtual 3D Teleportation in
  Real-time}. In \bibinfo{booktitle}{\emph{UIST}}.
\newblock


\bibitem[\protect\citeauthoryear{Pandey, Tkach, Yang, Pidlypenskyi, Taylor,
  Martin-Brualla, Tagliasacchi, Papandreou, Davidson, Keskin, Izadi, and
  Fanello}{Pandey et~al\mbox{.}}{2019}]%
        {simply_lookingood}
\bibfield{author}{\bibinfo{person}{Rohit Pandey}, \bibinfo{person}{Anastasia
  Tkach}, \bibinfo{person}{Shuoran Yang}, \bibinfo{person}{Pavel Pidlypenskyi},
  \bibinfo{person}{Jonathan Taylor}, \bibinfo{person}{Ricardo Martin-Brualla},
  \bibinfo{person}{Andrea Tagliasacchi}, \bibinfo{person}{George Papandreou},
  \bibinfo{person}{Philip Davidson}, \bibinfo{person}{Cem Keskin},
  \bibinfo{person}{Shahram Izadi}, {and} \bibinfo{person}{Sean Fanello}.}
  \bibinfo{year}{2019}\natexlab{}.
\newblock \showarticletitle{Volumetric Capture of Humans with a Single RGBD
  Camera via Semi-Parametric Learning}. In \bibinfo{booktitle}{\emph{CVPR}}.
\newblock


\bibitem[\protect\citeauthoryear{Pharr, Jakob, and Humphreys}{Pharr
  et~al\mbox{.}}{2016}]%
        {pharr}
\bibfield{author}{\bibinfo{person}{Matt Pharr}, \bibinfo{person}{Wenzel Jakob},
  {and} \bibinfo{person}{Greg Humphreys}.} \bibinfo{year}{2016}\natexlab{}.
\newblock \bibinfo{booktitle}{\emph{Physically Based Rendering: From Theory to
  Implementation} (\bibinfo{edition}{3rd} ed.)}.
\newblock \bibinfo{publisher}{Morgan Kaufmann Publishers Inc.}
\newblock


\bibitem[\protect\citeauthoryear{Ren, Dong, Lin, Tong, and Guo}{Ren
  et~al\mbox{.}}{2015}]%
        {ren_image_2015}
\bibfield{author}{\bibinfo{person}{Peiran Ren}, \bibinfo{person}{Yue Dong},
  \bibinfo{person}{Stephen Lin}, \bibinfo{person}{Xin Tong}, {and}
  \bibinfo{person}{Baining Guo}.} \bibinfo{year}{2015}\natexlab{}.
\newblock \showarticletitle{Image Based Relighting Using Neural Networks}.
\newblock \bibinfo{journal}{\emph{ACM TOG}} (\bibinfo{year}{2015}).
\newblock


\bibitem[\protect\citeauthoryear{Ronneberger, Fischer, and Brox}{Ronneberger
  et~al\mbox{.}}{2015}]%
        {ronneberger2015u}
\bibfield{author}{\bibinfo{person}{Olaf Ronneberger}, \bibinfo{person}{Philipp
  Fischer}, {and} \bibinfo{person}{Thomas Brox}.}
  \bibinfo{year}{2015}\natexlab{}.
\newblock \showarticletitle{U-net: Convolutional networks for biomedical image
  segmentation}. In \bibinfo{booktitle}{\emph{MICCAI}}.
\newblock


\bibitem[\protect\citeauthoryear{Saxena, Sun, and Ng}{Saxena
  et~al\mbox{.}}{2008}]%
        {saxena2008make3d}
\bibfield{author}{\bibinfo{person}{Ashutosh Saxena}, \bibinfo{person}{Min Sun},
  {and} \bibinfo{person}{Andrew~Y Ng}.} \bibinfo{year}{2008}\natexlab{}.
\newblock \showarticletitle{Make3d: Learning 3d scene structure from a single
  still image}.
\newblock \bibinfo{journal}{\emph{IEEE TPAMI}} (\bibinfo{year}{2008}).
\newblock


\bibitem[\protect\citeauthoryear{Sen, Chen, Garg, Marschner, Horowitz, Levoy,
  and Lensch}{Sen et~al\mbox{.}}{2005}]%
        {sen_dual_2005}
\bibfield{author}{\bibinfo{person}{Pradeep Sen}, \bibinfo{person}{Billy Chen},
  \bibinfo{person}{Gaurav Garg}, \bibinfo{person}{Stephen~R. Marschner},
  \bibinfo{person}{Mark Horowitz}, \bibinfo{person}{Marc Levoy}, {and}
  \bibinfo{person}{Hendrik P.~A. Lensch}.} \bibinfo{year}{2005}\natexlab{}.
\newblock \showarticletitle{Dual {Photography}}. In
  \bibinfo{booktitle}{\emph{SIGGRAPH}}.
\newblock


\bibitem[\protect\citeauthoryear{Sengupta, Jayaram, Curless, Seitz, and
  Kemelmacher-Shlizerman}{Sengupta et~al\mbox{.}}{2020}]%
        {BMSengupta20}
\bibfield{author}{\bibinfo{person}{Soumyadip Sengupta}, \bibinfo{person}{Vivek
  Jayaram}, \bibinfo{person}{Brian Curless}, \bibinfo{person}{Steve Seitz},
  {and} \bibinfo{person}{Ira Kemelmacher-Shlizerman}.}
  \bibinfo{year}{2020}\natexlab{}.
\newblock \showarticletitle{Background Matting: The World is Your Green
  Screen}. In \bibinfo{booktitle}{\emph{CVPR}}.
\newblock


\bibitem[\protect\citeauthoryear{Sengupta, Kanazawa, Castillo, and
  Jacobs}{Sengupta et~al\mbox{.}}{2018}]%
        {sfsnetSengupta18}
\bibfield{author}{\bibinfo{person}{Soumyadip Sengupta}, \bibinfo{person}{Angjoo
  Kanazawa}, \bibinfo{person}{Carlos~D. Castillo}, {and}
  \bibinfo{person}{David~W. Jacobs}.} \bibinfo{year}{2018}\natexlab{}.
\newblock \showarticletitle{SfSNet: Learning Shape, Refectance and Illuminance
  of Faces in the Wild}. In \bibinfo{booktitle}{\emph{CVPR}}.
\newblock


\bibitem[\protect\citeauthoryear{Shysheya, Zakharov, Aliev, Bashirov, Burkov,
  Iskakov, Ivakhnenko, Malkov, Pasechnik, Ulyanov, Vakhitov, and
  Lempitsky}{Shysheya et~al\mbox{.}}{2019}]%
        {shysheya_textured_nodate}
\bibfield{author}{\bibinfo{person}{Aliaksandra Shysheya}, \bibinfo{person}{Egor
  Zakharov}, \bibinfo{person}{Kara-Ali Aliev}, \bibinfo{person}{Renat
  Bashirov}, \bibinfo{person}{Egor Burkov}, \bibinfo{person}{Karim Iskakov},
  \bibinfo{person}{Aleksei Ivakhnenko}, \bibinfo{person}{Yury Malkov},
  \bibinfo{person}{Igor Pasechnik}, \bibinfo{person}{Dmitry Ulyanov},
  \bibinfo{person}{Alexander Vakhitov}, {and} \bibinfo{person}{Victor
  Lempitsky}.} \bibinfo{year}{2019}\natexlab{}.
\newblock \showarticletitle{Textured {Neural} {Avatars}}. In
  \bibinfo{booktitle}{\emph{CVPR}}.
\newblock


\bibitem[\protect\citeauthoryear{Simonyan and Zisserman}{Simonyan and
  Zisserman}{2014}]%
        {simonyan2014very}
\bibfield{author}{\bibinfo{person}{Karen Simonyan} {and}
  \bibinfo{person}{Andrew Zisserman}.} \bibinfo{year}{2014}\natexlab{}.
\newblock \showarticletitle{Very deep convolutional networks for large-scale
  image recognition}.
\newblock \bibinfo{journal}{\emph{arXiv preprint arXiv:1409.1556}}
  (\bibinfo{year}{2014}).
\newblock


\bibitem[\protect\citeauthoryear{Sitzmann, Thies, Heide, Nießner, Wetzstein,
  and Zollhöfer}{Sitzmann et~al\mbox{.}}{2019a}]%
        {sitzmann_deepvoxels:_2019}
\bibfield{author}{\bibinfo{person}{Vincent Sitzmann}, \bibinfo{person}{Justus
  Thies}, \bibinfo{person}{Felix Heide}, \bibinfo{person}{Matthias Nießner},
  \bibinfo{person}{Gordon Wetzstein}, {and} \bibinfo{person}{Michael
  Zollhöfer}.} \bibinfo{year}{2019}\natexlab{a}.
\newblock \showarticletitle{{DeepVoxels}: {Learning} {Persistent} 3D {Feature}
  {Embeddings}}. In \bibinfo{booktitle}{\emph{CVPR}}.
\newblock


\bibitem[\protect\citeauthoryear{Sitzmann, Zollh{\"o}fer, and
  Wetzstein}{Sitzmann et~al\mbox{.}}{2019b}]%
        {sitzmann2019scene}
\bibfield{author}{\bibinfo{person}{Vincent Sitzmann}, \bibinfo{person}{Michael
  Zollh{\"o}fer}, {and} \bibinfo{person}{Gordon Wetzstein}.}
  \bibinfo{year}{2019}\natexlab{b}.
\newblock \showarticletitle{Scene representation networks: Continuous
  3d-structure-aware neural scene representations}. In
  \bibinfo{booktitle}{\emph{Advances in Neural Information Processing
  Systems}}. \bibinfo{pages}{1121--1132}.
\newblock


\bibitem[\protect\citeauthoryear{Snavely, Seitz, and Szeliski}{Snavely
  et~al\mbox{.}}{2006}]%
        {snavely2006photo}
\bibfield{author}{\bibinfo{person}{Noah Snavely}, \bibinfo{person}{Steven~M
  Seitz}, {and} \bibinfo{person}{Richard Szeliski}.}
  \bibinfo{year}{2006}\natexlab{}.
\newblock \showarticletitle{Photo tourism: exploring photo collections in 3D}.
  In \bibinfo{booktitle}{\emph{SIGGRAPH}}.
\newblock


\bibitem[\protect\citeauthoryear{Srinivasan, Tucker, Barron, Ramamoorthi, Ng,
  and Snavely}{Srinivasan et~al\mbox{.}}{2019}]%
        {SrinivasanCVPR2019}
\bibfield{author}{\bibinfo{person}{Pratul~P. Srinivasan},
  \bibinfo{person}{Richard Tucker}, \bibinfo{person}{Jonathan~T. Barron},
  \bibinfo{person}{Ravi Ramamoorthi}, \bibinfo{person}{Ren Ng}, {and}
  \bibinfo{person}{Noah Snavely}.} \bibinfo{year}{2019}\natexlab{}.
\newblock \showarticletitle{Pushing the Boundaries of View Extrapolation with
  Multiplane Images}. In \bibinfo{booktitle}{\emph{CVPR}}.
\newblock


\bibitem[\protect\citeauthoryear{Sun, Barron, Tsai, Xu, Yu, Fyffe, Rhemann,
  Busch, Debevec, and Ramamoorthi}{Sun et~al\mbox{.}}{2019}]%
        {sun_single_2019}
\bibfield{author}{\bibinfo{person}{Tiancheng Sun}, \bibinfo{person}{Jonathan~T.
  Barron}, \bibinfo{person}{Yun{-}Ta Tsai}, \bibinfo{person}{Zexiang Xu},
  \bibinfo{person}{Xueming Yu}, \bibinfo{person}{Graham Fyffe},
  \bibinfo{person}{Christoph Rhemann}, \bibinfo{person}{Jay Busch},
  \bibinfo{person}{Paul~E. Debevec}, {and} \bibinfo{person}{Ravi Ramamoorthi}.}
  \bibinfo{year}{2019}\natexlab{}.
\newblock \showarticletitle{Single Image Portrait Relighting}. In
  \bibinfo{booktitle}{\emph{SIGGRAPH}}.
\newblock


\bibitem[\protect\citeauthoryear{Sun, Xu, Zhang, Fanello, Rhemann, Debevec,
  Tsai, Barron, and Ramamoorthi}{Sun et~al\mbox{.}}{2020}]%
        {sun2020light}
\bibfield{author}{\bibinfo{person}{Tiancheng Sun}, \bibinfo{person}{Zexiang
  Xu}, \bibinfo{person}{Xiuming Zhang}, \bibinfo{person}{Sean Fanello},
  \bibinfo{person}{Christoph Rhemann}, \bibinfo{person}{Paul Debevec},
  \bibinfo{person}{Yun-Ta Tsai}, \bibinfo{person}{Jonathan~T Barron}, {and}
  \bibinfo{person}{Ravi Ramamoorthi}.} \bibinfo{year}{2020}\natexlab{}.
\newblock \showarticletitle{Light Stage Super-Resolution: Continuous
  High-Frequency Relighting}.
\newblock \bibinfo{journal}{\emph{arXiv preprint arXiv:2010.08888}}
  (\bibinfo{year}{2020}).
\newblock


\bibitem[\protect\citeauthoryear{Tewari, Elgharib, Bharaj, Bernard, Seidel,
  P{\'e}rez, Z{\"o}llhofer, and Theobalt}{Tewari et~al\mbox{.}}{2020a}]%
        {tewari2020stylerig}
\bibfield{author}{\bibinfo{person}{Ayush Tewari}, \bibinfo{person}{Mohamed
  Elgharib}, \bibinfo{person}{Gaurav Bharaj}, \bibinfo{person}{Florian
  Bernard}, \bibinfo{person}{Hans-Peter Seidel}, \bibinfo{person}{Patrick
  P{\'e}rez}, \bibinfo{person}{Michael Z{\"o}llhofer}, {and}
  \bibinfo{person}{Christian Theobalt}.} \bibinfo{year}{2020}\natexlab{a}.
\newblock \showarticletitle{StyleRig: Rigging StyleGAN for 3D Control over
  Portrait Images, CVPR 2020}. In \bibinfo{booktitle}{\emph{CVPR}}.
\newblock


\bibitem[\protect\citeauthoryear{Tewari, Fried, Thies, Sitzmann, Lombardi,
  Sunkavalli, Martin-Brualla, Simon, Saragih, Nießner, Pandey, Fanello,
  Wetzstein, Zhu, Theobalt, Agrawala, Shechtman, Goldman, and
  Zollhoefer}{Tewari et~al\mbox{.}}{2020b}]%
        {tewari2020state}
\bibfield{author}{\bibinfo{person}{Ayush Tewari}, \bibinfo{person}{Ohad Fried},
  \bibinfo{person}{Justus Thies}, \bibinfo{person}{Vincent Sitzmann},
  \bibinfo{person}{Stephen Lombardi}, \bibinfo{person}{Kalyan Sunkavalli},
  \bibinfo{person}{Ricardo Martin-Brualla}, \bibinfo{person}{Tomas Simon},
  \bibinfo{person}{Jason Saragih}, \bibinfo{person}{Matthias Nießner},
  \bibinfo{person}{Rohit Pandey}, \bibinfo{person}{Sean Fanello},
  \bibinfo{person}{Gordon Wetzstein}, \bibinfo{person}{Jun-Yan Zhu},
  \bibinfo{person}{Christian Theobalt}, \bibinfo{person}{Maneesh Agrawala},
  \bibinfo{person}{Eli Shechtman}, \bibinfo{person}{Dan~B Goldman}, {and}
  \bibinfo{person}{Michael Zollhoefer}.} \bibinfo{year}{2020}\natexlab{b}.
\newblock \showarticletitle{State of the Art on Neural Rendering}. In
  \bibinfo{booktitle}{\emph{Eurographics}}.
\newblock


\bibitem[\protect\citeauthoryear{Thies, Zollh{\"o}fer, Theobalt, Stamminger,
  and Nie{\ss}ner}{Thies et~al\mbox{.}}{2020}]%
        {thies2020image}
\bibfield{author}{\bibinfo{person}{Justus Thies}, \bibinfo{person}{Michael
  Zollh{\"o}fer}, \bibinfo{person}{Christian Theobalt}, \bibinfo{person}{Marc
  Stamminger}, {and} \bibinfo{person}{Matthias Nie{\ss}ner}.}
  \bibinfo{year}{2020}\natexlab{}.
\newblock \showarticletitle{Image-guided neural object rendering}. In
  \bibinfo{booktitle}{\emph{International Conference on Learning
  Representations}}.
\newblock


\bibitem[\protect\citeauthoryear{Thies, Zollhöfer, and Nießner}{Thies
  et~al\mbox{.}}{2019}]%
        {thies_deferred_2019}
\bibfield{author}{\bibinfo{person}{Justus Thies}, \bibinfo{person}{Michael
  Zollhöfer}, {and} \bibinfo{person}{Matthias Nießner}.}
  \bibinfo{year}{2019}\natexlab{}.
\newblock \showarticletitle{Deferred neural rendering: image synthesis using
  neural textures}. In \bibinfo{booktitle}{\emph{SIGGRAPH}}.
\newblock


\bibitem[\protect\citeauthoryear{Wang, Bovik, Sheikh, and Simoncelli}{Wang
  et~al\mbox{.}}{2004}]%
        {zhou_wang_image_2004}
\bibfield{author}{\bibinfo{person}{Zhou Wang}, \bibinfo{person}{Alan~C. Bovik},
  \bibinfo{person}{Hamid~R. Sheikh}, {and} \bibinfo{person}{Eero~P.
  Simoncelli}.} \bibinfo{year}{2004}\natexlab{}.
\newblock \showarticletitle{Image quality assessment: from error visibility to
  structural similarity}.
\newblock \bibinfo{journal}{\emph{IEEE TIP}} (\bibinfo{year}{2004}).
\newblock


\bibitem[\protect\citeauthoryear{Wiles, Gkioxari, Szeliski, and Johnson}{Wiles
  et~al\mbox{.}}{2020}]%
        {wiles2020synsin}
\bibfield{author}{\bibinfo{person}{Olivia Wiles}, \bibinfo{person}{Georgia
  Gkioxari}, \bibinfo{person}{Richard Szeliski}, {and} \bibinfo{person}{Justin
  Johnson}.} \bibinfo{year}{2020}\natexlab{}.
\newblock \showarticletitle{Synsin: End-to-end view synthesis from a single
  image}. In \bibinfo{booktitle}{\emph{Proceedings of the IEEE/CVF Conference
  on Computer Vision and Pattern Recognition}}. \bibinfo{pages}{7467--7477}.
\newblock


\bibitem[\protect\citeauthoryear{Woodham}{Woodham}{1980}]%
        {photometric_stereo}
\bibfield{author}{\bibinfo{person}{Robert~J. Woodham}.}
  \bibinfo{year}{1980}\natexlab{}.
\newblock \showarticletitle{{Photometric Method For Determining Surface
  Orientation From Multiple Images}}.
\newblock \bibinfo{journal}{\emph{Optical Engineering}} (\bibinfo{year}{1980}).
\newblock


\bibitem[\protect\citeauthoryear{Xu, Bi, Sunkavalli, Hadap, Su, and
  Ramamoorthi}{Xu et~al\mbox{.}}{2019}]%
        {xu_deep_2019}
\bibfield{author}{\bibinfo{person}{Zexiang Xu}, \bibinfo{person}{Sai Bi},
  \bibinfo{person}{Kalyan Sunkavalli}, \bibinfo{person}{Sunil Hadap},
  \bibinfo{person}{Hao Su}, {and} \bibinfo{person}{Ravi Ramamoorthi}.}
  \bibinfo{year}{2019}\natexlab{}.
\newblock \showarticletitle{Deep view synthesis from sparse photometric
  images}. In \bibinfo{booktitle}{\emph{SIGGRAPH}}.
\newblock


\bibitem[\protect\citeauthoryear{Xu, Sunkavalli, Hadap, and Ramamoorthi}{Xu
  et~al\mbox{.}}{2018}]%
        {xu_deep_2018}
\bibfield{author}{\bibinfo{person}{Zexiang Xu}, \bibinfo{person}{Kalyan
  Sunkavalli}, \bibinfo{person}{Sunil Hadap}, {and} \bibinfo{person}{Ravi
  Ramamoorthi}.} \bibinfo{year}{2018}\natexlab{}.
\newblock \showarticletitle{Deep image-based relighting from optimal sparse
  samples}.
\newblock \bibinfo{journal}{\emph{SIGGRAPH 2018}}.
\newblock


\bibitem[\protect\citeauthoryear{Zhang, Isola, Efros, Shechtman, and
  Wang}{Zhang et~al\mbox{.}}{2018}]%
        {zhang_unreasonable_2018}
\bibfield{author}{\bibinfo{person}{Richard Zhang}, \bibinfo{person}{Phillip
  Isola}, \bibinfo{person}{Alexei~A. Efros}, \bibinfo{person}{Eli Shechtman},
  {and} \bibinfo{person}{Oliver Wang}.} \bibinfo{year}{2018}\natexlab{}.
\newblock \showarticletitle{The {Unreasonable} {Effectiveness} of {Deep}
  {Features} as a {Perceptual} {Metric}}. In \bibinfo{booktitle}{\emph{CVPR}}.
\newblock


\bibitem[\protect\citeauthoryear{Zhang, Barron, Tsai, Pandey, Zhang, Ng, and
  Jacobs}{Zhang et~al\mbox{.}}{2020}]%
        {zhang2020portrait}
\bibfield{author}{\bibinfo{person}{Xuaner Zhang}, \bibinfo{person}{Jonathan~T.
  Barron}, \bibinfo{person}{Yun-Ta Tsai}, \bibinfo{person}{Rohit Pandey},
  \bibinfo{person}{Xiuming Zhang}, \bibinfo{person}{Ren Ng}, {and}
  \bibinfo{person}{David~E. Jacobs}.} \bibinfo{year}{2020}\natexlab{}.
\newblock \showarticletitle{Portrait Shadow Manipulation}.
\newblock \bibinfo{journal}{\emph{ACM Transactions on Graphics (TOG)}}
  \bibinfo{volume}{39}, \bibinfo{number}{4}.
\newblock


\bibitem[\protect\citeauthoryear{Zhou, Tucker, Flynn, Fyffe, and Snavely}{Zhou
  et~al\mbox{.}}{2018}]%
        {zhou_stereo_2018}
\bibfield{author}{\bibinfo{person}{Tinghui Zhou}, \bibinfo{person}{Richard
  Tucker}, \bibinfo{person}{John Flynn}, \bibinfo{person}{Graham Fyffe}, {and}
  \bibinfo{person}{Noah Snavely}.} \bibinfo{year}{2018}\natexlab{}.
\newblock \showarticletitle{Stereo magnification: learning view synthesis using
  multiplane images}. In \bibinfo{booktitle}{\emph{SIGGRAPH}}.
\newblock


\bibitem[\protect\citeauthoryear{{Zickler}, {Ramamoorthi}, {Enrique}, and
  {Belhumeur}}{{Zickler} et~al\mbox{.}}{2006}]%
        {zickler06}
\bibfield{author}{\bibinfo{person}{T. {Zickler}}, \bibinfo{person}{R.
  {Ramamoorthi}}, \bibinfo{person}{S. {Enrique}}, {and} \bibinfo{person}{P.~N.
  {Belhumeur}}.} \bibinfo{year}{2006}\natexlab{}.
\newblock \showarticletitle{Reflectance sharing: predicting appearance from a
  sparse set of images of a known shape}.
\newblock \bibinfo{journal}{\emph{IEEE TPAMI}} (\bibinfo{year}{2006}).
\newblock


\end{thebibliography}

\end{document}